\RequirePackage{fix-cm}
\documentclass[twocolumn]{svjour3}          % twocolumn
\smartqed  % flush right qed marks, e.g. at end of proof

\usepackage{array}
\usepackage{bm}
\usepackage{url}
\usepackage{todonotes}
\usepackage{setspace}
\usepackage{natbib}
\usepackage{graphicx}
\usepackage{multirow}
\usepackage{amsmath,amssymb,amsfonts}%
\usepackage{mathrsfs}%
\usepackage[title]{appendix}%
\usepackage{xcolor}%
\usepackage{textcomp}%
\usepackage{manyfoot}%
\usepackage{booktabs}%
\usepackage{algorithm}%
\usepackage{algpseudocode}%
\usepackage{listings}%
\usepackage{colortbl}
\usepackage{comment}
\usepackage{tabularx}

\usepackage[colorlinks=true]{hyperref}
\newcommand{\etal}{\textit{et al.}}
\newcommand{\ie}{\textit{i.e.}}
\newcommand{\eg}{\textit{e.g.}}

\begin{document}
\sloppy

\title{FusionBooster: A Unified Image Fusion Boosting Paradigm}



\author{Chunyang~Cheng      \and
        Tianyang~Xu         \and
        Xiao-Jun~Wu         \and
        Hui~Li              \and        
        Xi~Li       \and
        Josef~Kittler
}
%\author[1]{\fnm{Chunyang} \sur{Cheng}}\email{chunyang\_cheng@163.com}


\institute{C.~Cheng, T.~Xu, X. J.~Wu*, and H.~Li \at
              School of Artificial Intelligence and Computer Science\\
              Jiangnan University, Wuxi, 214122, China. \\
              *\email{wu\_xiaojun@jiangnan.edu.cn}           %  \\
           \and
              X.~Li \at
              College of Computer Science and Technology\\
              Zhejiang University, Hangzhou, 310027, China. \\
              \email{xilizju@zju.edu.cn}
           \and
              J.~Kitter \at
              Centre for Vision, Speech and Signal Processing\\
              University of Surrey, Guildford, GU2 7XH, UK. \\
              \email{j.kittler@surrey.ac.uk}
}

\date{Received: July, 2023 / Accepted: date}

\maketitle

\begin{abstract}
In recent years, numerous ideas have emerged for designing a mutually reinforcing mechanism or extra stages for the image fusion task, ignoring the inevitable gaps between different vision tasks and the computational burden.
%in the image fusion field.
%Essentially, all the existing formulations try to manage the diverse levels of information communicated by the source images to achieve the best fusion result. 
%We argue that there is a scope for improving the performance of existing methods further with the help of FusionBooster, a model specifically designed for the fusion task.
We argue that there is a scope to improve the fusion performance with the help of the FusionBooster, a model specifically designed for the fusion task.
%a fusion guidance method proposed in this paper.  
%lightweight 
In particular, our booster is based on the divide-and-conquer strategy controlled by an information probe.
The booster is composed of three building blocks: the probe units, the booster layer, and the assembling module. 
Given the result produced by a backbone method, the probe units assess the fused image and divide the results according to their information content.
This is instrumental in identifying missing information, as a step to its recovery.
The recovery of the degraded components along with the fusion guidance are the role of the booster layer.
Lastly, the assembling module is responsible for piecing these advanced components together to deliver the output.
We use concise reconstruction loss functions in conjunction with lightweight autoencoder models to formulate the learning task, with marginal computational complexity increase.
The experimental results obtained in various fusion tasks, as well as downstream detection tasks, consistently demonstrate that the proposed FusionBooster significantly improves the performance.
Our code will be publicly available at \url{https://github.com/AWCXV/FusionBooster}.

\end{abstract}

\section{Introduction}\label{sec1}

Image fusion is a technique aiming to combine complementary information from diverse modalities, or images with different shooting settings, into a single image.
The fused image, which becomes more informative, is expected to have enhanced visual quality, as well as boost the performance of downstream vision tasks.
This technique has been widely applied to different areas, including video surveillance, object tracking, remote sensing imaging, and medical diagnosis~\citep{xu2022mefsurvey,xu2019joint,zhang2021deepMFsurvey,tang2023exploring}.

Broadly speaking, the current image fusion tasks fall into two main categories, $i.e.$, multi-modal image fusion and digital photography fusion.
For instance, the infrared and visible image fusion (IVIF) task, which belongs to the former category, arises in many practical applications.
It aims to combine the rich scene texture from the visible image, with the robust thermal and structural information tapped from the infrared modality.
Since the infrared modality is insensitive to variations in the environmental condition, combining these complementary sources of information helps to enhance the visualization of challenging scenes, \textit{e.g.}, in the foggy or low-light environments~\citep{sun2022detfusion}.
On the other hand, multi-exposure image fusion (MEIF) and multi-focus image fusion (MFIF)  belong to the latter category (digital photography).
Specifically, the MEIF task is to combine the input overexposed and underexposed images in order to generate fusion results with an appropriate exposure setting~\citep{xu2020mef}.
The goal of the MFIF task is to produce a fully focused image by combining the near-focused and far-focused images at the input to counteract the depth-of-field limitation in imaging~\citep{zhang2021mffgan}.

\begin{figure}[t]
\begin{center}
\includegraphics[width=1\linewidth]{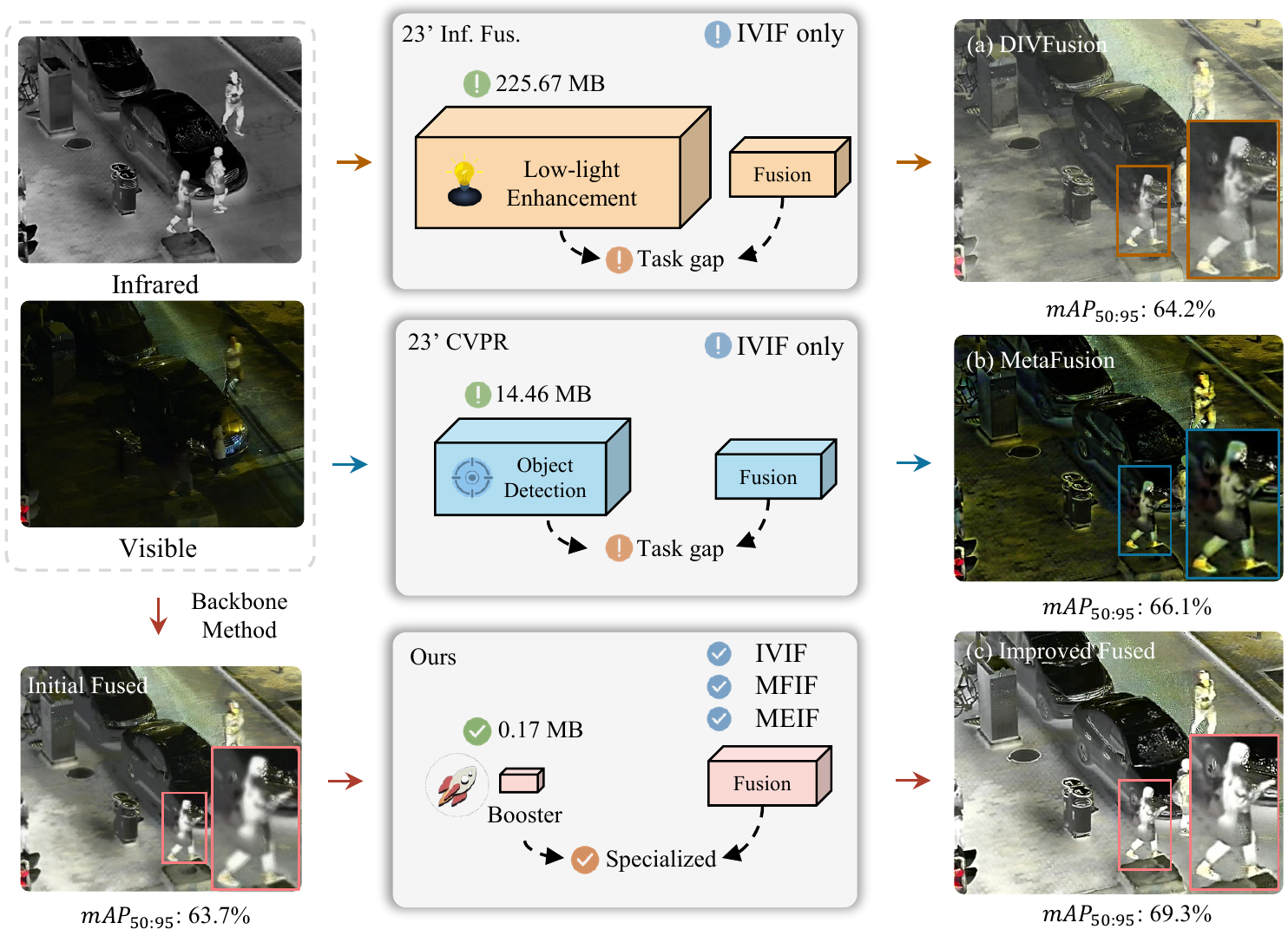}
\end{center}
   \caption{
   Comparison of the proposed FusionBooster and other advanced methods that contain additional enhancement models. The current algorithms are suffering from the issues of expensive computational cost, task gap and the lack of generalization ability. (Backbone method: DDcGAN~\citep{ma2020ddcgan})
}
\label{figureMotivation}
\end{figure}

%Since the infrared image are not sensitive to the environment discrepancy, the fused images can provide robust information in the terrible environment, \textit{e.g.}, in the rainy day and dark environment.

In the primary exploration stage, various signal processing techniques had been applied to accomplish the fusion process in the conventional paradigm exemplified by~\citep{ma2016gtf,liu2016csr,yang2018TLER,li2020mdlatlrr,li2020fastMEFDecomposition,chen2021multiTradition}.
However, the limitations of the classical feature extraction and fusion techniques motivated the emergence of deep learning-based fusion methods~\citep{li2018densefuse,xu2020u2fusion,zhang2021sdnet,tang2022piafusion,cheng2023mufusion}.
Currently, the trend has shifted towards the focus on the interplay between fusion and other vision tasks~\citep{huang2022reconet,tang2023divfusion,tang2022SeAFusion,xu2022rfnet}.
%A few studies also argue for the adoption of extra training stage in the fusion task~\citep{fu2021ppt,ma2022swinfusion,rao2022tgfuse}.
A few studies also argue for adopting an extra training stage in the fusion task~\citep{li2021rfn,zhao2023cddfuse}.
However, as shown in Fig.~\ref{figureMotivation}, the performance of current mainstream fusion methods is  highly impeded by three factors: the expensive computational overhead, the task gap, and the inadequate generalization ability.
%\begin{enumerate}
    %\item The expensive computational overhead.
    %\item The task gap between the image fusion and other vision tasks.
    %\item The generalization ability of the auxiliary models.
%\end{enumerate}

Specifically, the additional computational cost arises mainly from the other vision models or extra training stages incorporated in their methods.
Such expensive overhead can hinder the practical adaptation of the fusion algorithm to new scenes, when limited computational resources are available.
Furthermore, the introduction of other vision tasks also brings up the task gap issue in the current image fusion paradigm.
Typically, these methods disregard the potential discrepancy between the nature of information processing at low-level fusion, and high-level vision problems.
Consequently, the feedback from certain vision tasks may be completely inappropriate for the task of refining the fusion model.
That is, combining fusion and other vision tasks with different objectives may result in suboptimal fused images.
For example, as illustrated in (a) of Fig.~\ref{figureMotivation}, the introduction of the low-light enhancement model effectively improves the visible component of the fused images, but it is not very effective at maintaining thermal radiation information (the fusion task).
Similarly, the result (b) also indicates that the compatibility of the fusion and object detection tasks is not quite satisfactory, as the visualization effect is not promising and the detection precision is not significantly improved with the help of the detection model.
Finally, note that the current enhancement-based image fusion methods can only work in a specific fusion task.
The digital photography fusion tasks, \ie, the MFIF and the MEIF tasks, can not benefit from these paradigms, which demonstrates their deficiency in the generalization ability.

\begin{figure}[t]
\begin{center}
\includegraphics[width=0.7\linewidth]{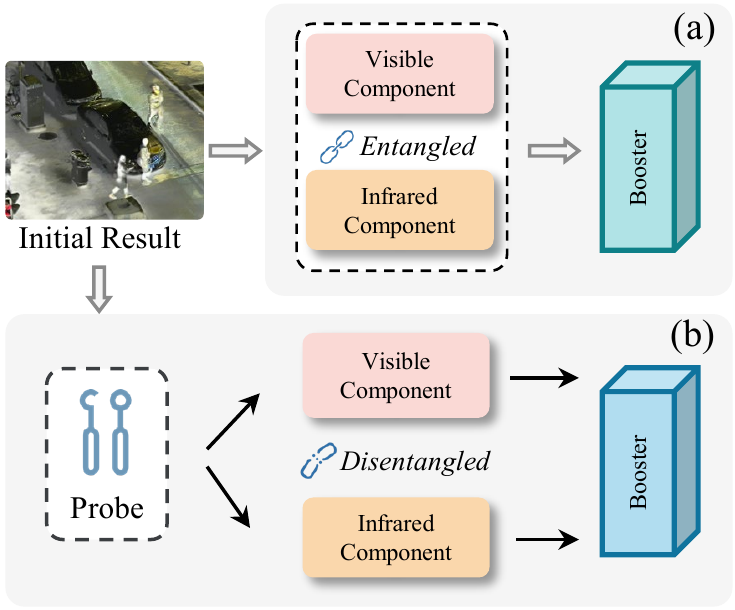}
\end{center}
   \caption{
   A comparison of the proposed divide and conquer boosting paradigm (b) and existing methods (a) relying on the booster (other vision models).
   The disentangled components allow us to better improve the fusion results in a fine-grained manner, which also provides us with the flexibility to handle more tasks, depending on the content.
}
\label{fig_information_probe}
\end{figure}

In this work, we propose the FusionBooster model to address the above-mentioned problems.
%, , we propose to boost the image fusion performance by considering the essence of the fusion task, \textit{i.e.}, the need for the preservation of information conveyed by the source images.
%Specifically, we use an information probe to gauge the quality of information conveyed by the source images in different components of the initial fusion result (Fig.~\ref{fig_information_probe} (b)).
%Due to the inappropriate assessment of the quality of information conveyed by the source images, or noise introduced in the first stage of processing, a good quality reconstruction of the source input from the fused image is generally impossible and the constituent components tend to degrade.
%Interestingly, the degree of degradation is correlated with the quality of the initial fusion results.
%Motivated by this observation, we incorporate, into the fusion system, a novel mechanism, which guides the process of reassembling these components to produce the fused image. 
%The mechanism enables delivering a better quality and more robust fusion result.
Firstly, our network only consists of several convolutional layers to formulate the encoder and decoder parts, forming a \textit{compact} model.
This design can effectively alleviate the expensive computational cost issue of existing enhancement-based methods.

Secondly, as our booster design reflects the characteristics of the image fusion task, we do not require additional vision models to intervene in the training process, thus avoiding the task gap issue.
As shown in Fig.~\ref{fig_information_probe} (b), given the initial fusion result, we specifically design an information \textit{probe} to reconstruct source images from it.
If the assessment of the information conveyed by the source image is of low quality, or the noise is introduced in the first stage of processing, a satisfied quality reconstruction of the source input from the fused image is generally impossible and the constituent components tend to degrade.
Interestingly, the degree of degradation is correlated with the quality of the initial fusion results (Section~\ref{analysisInformationProbe}).
Motivated by this observation, we incorporate, into the fusion system, a novel mechanism (\textit{booster}), which guides the process of reassembling these components to produce the fused image.
The mechanism enables the delivery of fusion results which are more robust and of better quality.
Compared with the existing methods (Fig.~\ref{fig_information_probe} (a)), our FusionBooster succeeds in the disentanglement of the fusion task and improves the fusion performance in a fine-grained manner.
%Thus, we use the divide-and-conquer strategy to boost the degraded components, instead of using the entangled manner adopted in most of the enhancement-based methods.

Thirdly, note that, depending on the characteristics of the fusion task, the output of the information probes is different for, \eg, the overexposed and underexposed components of the MEIF task.
This content-specific focus allows us to apply general operations on these detached components to benefit a series of fusion tasks, which alleviates the lack of generalization in the existing enhancement-based methods.
More specifically, as our probe can be regarded as a tool to gauge the information conveyed from the source images (\eg, from the infrared and visible images) into the fusion results, we design the corresponding booster layers to increase the information contained in these separate components.
In addition to this universal operation, we note that, in some studies, the experimental analysis has shown that the salient texture details can improve the performance of downstream vision tasks, as well as produce visually pleasing fused images~\citep{liu2022target,cheng2023mufusion}.
We argue that such enhancements can also consistently benefit different fusion tasks.
Thus, we take these findings into account in the design of the booster layers.
As a result, the upgraded methods can produce fused images that are more robust and simultaneously preserve the significant information from the source input to improve the performance of downstream tasks.

The contributions of this work can be summarized as follows:
\begin{itemize}
\item We devise an image fusion booster by analysing the quality of the initial fusion results by means of a dedicated \textit{Information Probe}.
\item In a new divide-and-conquer image fusion paradigm, the results of the analysis performed by the \textit{Information Probe} guide the refinement of the fused image with the help of a nested autoencoder network.

\item The proposed FusionBooster is a general enhancer, which can be applied to various image fusion methods, $e.g.$, traditional or learning-based algorithms, irrespective of the type of fusion task.
%(IVIF, MFIF, and MEIF).

\item The experimental results demonstrate that the proposed FusionBooster, in general, significantly enhances the performance of the state-of-the-art (SOTA) fusion methods and downstream detection tasks, with only a slight increase in the computational overhead. 
%Moreover, the improved fusion results yield the best performance in terms of the downstream detection task.

%, as well as the essence of the fusion process.
%adaption to the fusion task
%Owing to the substantial discrepancy between the fusion and other vision tasks, it is difficult to effectively improve the fusion quality.
%As shown in the pipeline of our FusionBooster, we further design the second stage to enhance the performance of fusion results from the first stage.
\end{itemize}

\section{Related Work}

\subsection{Learning-based Image Fusion Methods}
\label{related_work_learning}
In recent years, various learning-based image fusion methods have been proposed.
These methods can be roughly divided into three categories, $i.e.$, algorithms based on the generative adversarial networks (GAN), the autoencoders (AE), and the regular convolutional neural networks (CNN).
Specifically, the GAN-based methods rely on the adversarial game established between the generator and the discriminator to produce the fusion results~\citep{fu2021image,ma2020ganmcc}.
A representative work is the DDcGAN proposed by Ma~\etal~\citep{ma2020ddcgan}, which uses two discriminators to enable the fused images to preserve the useful information from the infrared and visible images.
\begin{figure}[t]
\begin{center}
\includegraphics[width=1\linewidth]{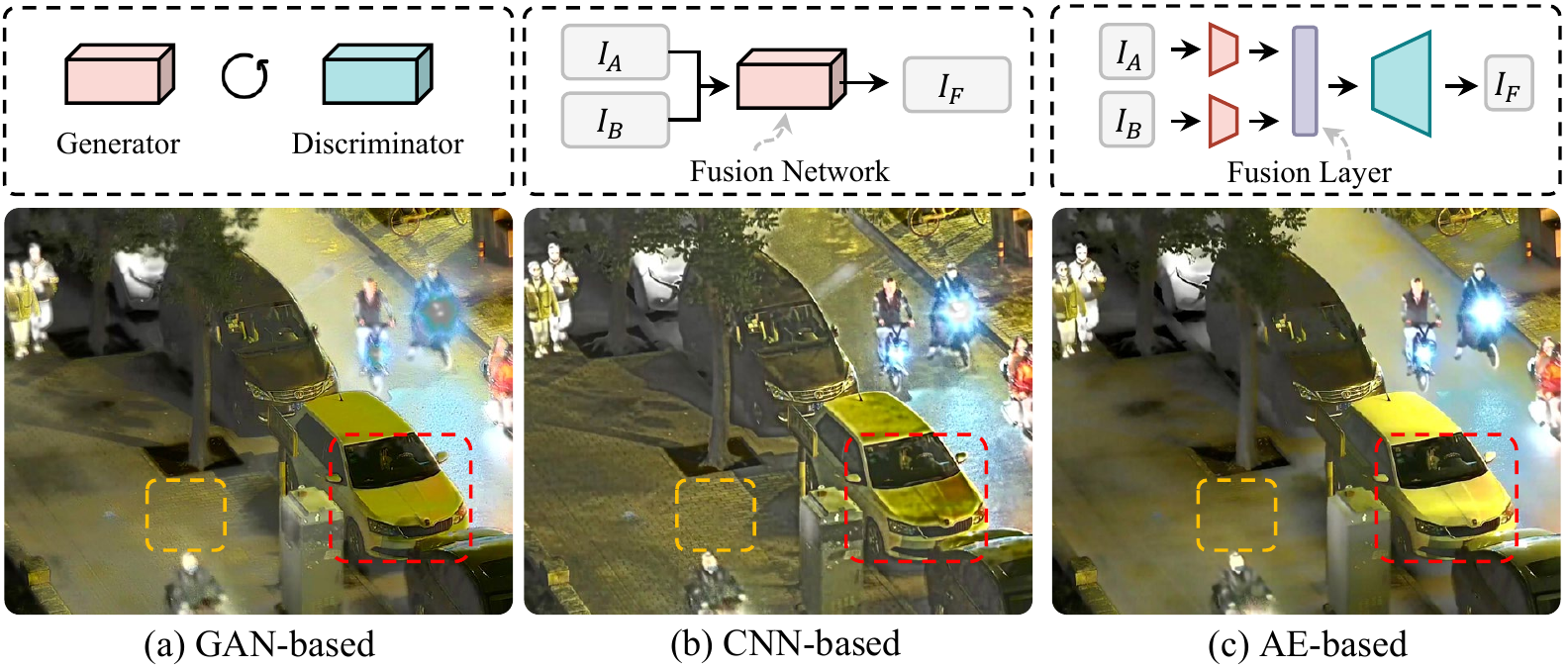}
\end{center}
   \caption{
   A comparison of different learning-based image fusion methods.
   The AE-based method, DenseFuse~\citep{li2018densefuse}  suffers from a bias issue, by biasing toward the infrared modality, which leads to the information loss in the fusion result (yellow boxes).
   However, as denoted by the red boxes, the AE-based method can produce more visually pleasing fused images, compared with the other two paradigms (DDcGAN and MUFusion~\citep{cheng2023mufusion}).
}
\label{fig_comparison_leaning}
\end{figure}
However, according to the investigations in~\citep{ma2019fusiongan,xu2020mef,rao2023tgfuse}, noise and artifacts are also incorporated into the fusion result, as part of the adversarial learning (pedestrians in Fig.~\ref{fig_comparison_leaning} (a)).
%Besides, the training proportion of the generator and discriminator also requires handcrafted adaption according to the speci.

In the MEIF field, taking into consideration the structural similarity, Prabhakar \textit{et al.} use an autoencoder to integrate the information from underexposed and overexposed images~\citep{prabhakar2017deepfuse}.
Li and Wu extend its application to the IVIF task~\citep{li2018densefuse} and a series of AE-based algorithms are proposed in~\citep{li2020nestfuse,fu2021dualbranchauto,li2021rfn}.
Although the authors devise elaborate fusion rules and even utilize a trainable network to learn an optimal fusion strategy, the bias issues are  still encountered in these methods, which leads to the loss of information (the ground in Fig.~\ref{fig_comparison_leaning} (c)).
%To handle this problem, we utilize a booster layer to respectively fix the degraded (caused by the inadequate information) components detached from the fusion results.
%and it dose not involve any mixing operations of the source images. 
%(Fig.~\ref{figureOverallPipeline}). 
%the dynamic balance between the generator and discriminator only contributes to the fusion results slightly and the quality of the fused images is not promising.
On the other hand, for the CNN-based methods~\citep{zhang2020rethinking,zhang2020ifcnn,long2021rxdnfuse,cheng2023mufusion}, they eliminate the handcrafted feature aggregation processes.
%output of these methods are highly dependent on the loss function used to optimize the fusion network.
%These end-to-end methods eliminate handcrafted feature aggregation processes.
However, the loss functions used in these approaches still rely on the empirical design based on the information theory~\citep{xu2020u2fusion}, activity level maps~\citep{cheng2023mufusion} or some choose-max strategies~\citep{zhang2021sdnet,tang2022SeAFusion}, which share similar risks with the handcrafted fusion rule designs.

In general, CNN-based methods and  GAN-based methods usually mix the feature extraction and feature aggregation processes up.
In contrast, in the AE-based methods, these two stages are separated.
Consequently, although the fusion layer design can sometimes give rise to  information loss, the fused images of the AE-based methods usually are more pleasing, compared with the aforementioned two paradigms (red boxes of Fig.~\ref{fig_comparison_leaning}).
Considering this merit of the AE-based approach, we adopt this paradigm in our FusionBooster and propose a nested AE network to first perceive and then reconstruct the initial fusion result.
As depicted in Fig.~\ref{figureMotivation}, by virtue of the FusionBooster, most of the noise and artifacts contained in the fused images can effectively be eliminated.
Note that, the fusion focus of the backbone method is retained in the enhanced result, \textit{e.g.}, both the salient thermal information and the rich texture details are preserved. 

%As a result, the image quality of these CNN-based methods usually cannot catch that of the AE-based method, which has been pre-trained on large amount dataset.

%As shown in red highlighted regions of the Fig.~\ref{fig_comparison_leaning} (b), the image quality of the CNN-based method cannot catch that of the AE-based method.

%However, even well-designed loss functions cannot avoid a bias issue, which results in suboptimal output.
%The image quality related loss function designs instead become the shortcomings of these methods.     

%More importantly, without an extra stage to fine-tune the fused images, these approaches basically generate the output in one shot.
%Our FusionBooster is specifically designed to address this problem. It uses a second stage to refine the initial  results by removing the artifacts, adjusting the exposure setting, or sharpening the edge information.
%Only minor computational costs will be introduced in the backbone models when using this booster.
%In this paper, we demonstrate that the reconstructed source images and an agent model can reproduce their results with higher image quality.
%Besides, those image quality loss functions 
%the loss function is actually 
%While our FusionBooster can further reduce the lost information and improve the overall image quality based on the fusion results of stage one.

\begin{figure*}[t]
\begin{center}
\includegraphics[width=1\linewidth]{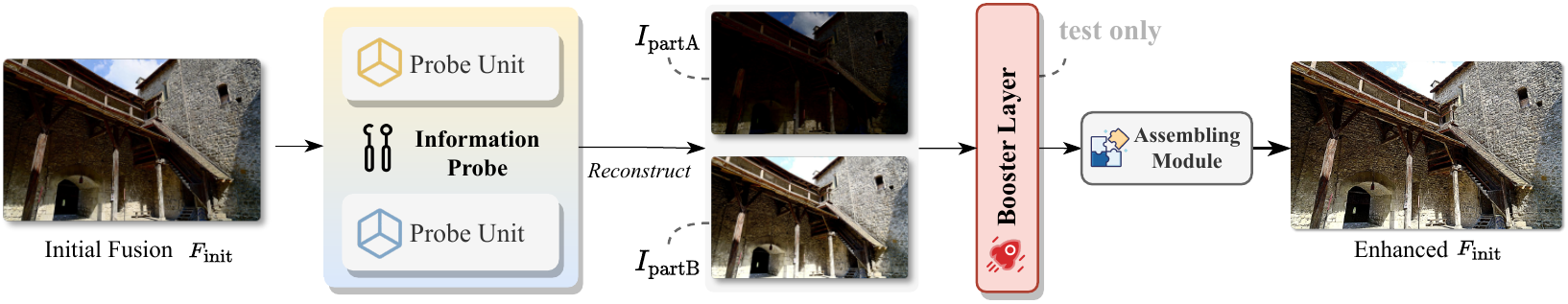}
\end{center}
   \caption{The pipeline of the proposed FusionBooster for the MEIF task (Backbone: U2Fusion).
   Our booster is composed of three parts, \textit{i.e.}, the information probe, the booster layer, and the assembling (ASE) module.
   %The probe unit learns to decompose the initial result of stage one into source components $I_\textrm{partA}$ and $I_\textrm{partB}$ by utilizing the perception loss functions.
   The information probe first perceives the source components $I_\textrm{partA}$ and $I_\textrm{partB}$ in the initial result.
   The ASE module will piece these components together to rebuild the initial result.
   In the test phase, the degraded components are fine-tuned in the booster layer and the ASE module correspondingly yields the enhanced result.
   %This process will be achieved by optimizing two probe networks (ProNet) with the loss functions of $Loss_{\textrm{divA}}$ and $Loss_{\textrm{divB}}$.
   %After that, the ASE module will regard these two components as input, with the recovery loss being used to train the reconstruction of $F_{\textrm{init}}$.
   }
\label{figureOverallPipeline}
\end{figure*}

\subsection{Image Fusion Methods with Integrated Enhancement Models}
Some of the image fusion methods are derived from the CNN-based approaches.
The significant difference lies in the use of an additional vision model or some other complementary stages to enhance the fusion performance~\citep{li2021rfn,liu2022target}.
%These additional models are expected to boost the fusion performance.
%Currently, researchers have realized the limitation of existing image fusion methods and are seeking new research directions on this topic.
%In the image fusion field, a popular research trend is the integration  of image fusion tasks and vision problems.
Specifically, the methods combined with other vision tasks train the fusion model and the detection or segmentation model in a joint or mutually reinforcing manner~\citep{sun2022detfusion,tang2022SeAFusion}.
In this way, the performance of both the related vision task and the IVIF task is expected to benefit.
In MetaFusion~\citep{zhao2023metafusion}, Zhao~\etal~address the task gap issue of these methods and propose to use a meta-feature embedding from the detection model to alleviate it.
However, their attempt is not completely satisfactory, as the combination of these features does not deliver robust fused images (result (b) in Fig.~\ref{figureMotivation}).

In contrast, the methods with an additional stage to learn the feature aggregation process do not suffer from the task gap issue.
In RFN-Nest~\citep{li2021rfn}, Li~\etal~replace the fusion layer from an AE-based method with a learnable fusion network.
The image fusion task is now transferred into the feature aggregation task.
However, this transformation does not disentangle the image fusion tasks effectively, as the fusion of the feature maps is as tricky as the fusion at the pixel level.
Consequently, with additional end-to-end requirements, the quality of the fused images of the RFN-Nest cannot catch that of the traditional AE-based methods~\citep{cheng2023mufusion}.

%However, although these methods try to take advantage of the AE-based methods, the loss function used in the second stage will take dominant position and impede the fusion performance.
%In fact, this methods can be attributed to the aforementioned CNN-based methods, except that the methods will receive the feature maps input and fuse the features instead of the source images.
%Our Fusion Booster directly deal with the fusion results and gauge the missing information. The backbone method already provide coarse-grained aggregation of source images, we can borrow the AE-based structure to further enhance them.
%Although producing promising results, the vision task incurs expensive annotation costs, which precludes its wide adoption in practical scenarios.
%Another weakness of these methods is that the feedback from other tasks is constrained by the semantic gap between various problems.
To address the above-mentioned issues of the existing enhancement-based methods, we design an information probe module for the image fusion task.
The role of this module is to sense the missing information in the initial fusion result.
In this way, we change the fusion enhancement objective to that of recovery of the source components by this module, which is more feasible.
Besides, as our FusionBooster does not require a joint training scheme, it can be even used to enhance the performance of traditional image fusion approaches.
Although no additional vision model is adopted in our booster, the experimental results demonstrate that, the information preservation strategies and the sharpening technique used in our booster can also significantly upgrade the performance of downstream detection tasks.

%exploit the inner properties of the fusion result by dividing each part of the source images from it.
%This transformation between the fusion modality and source inputs are not reversible, which results in degraded images.
%Based on this observation, we indirectly enhance the fusion results by fixing the middle degraded images, making the subsequent reconstruction process smoother and yield more robust output.
%Furthermore, the 
%We demonstrate that in this way we can simultaneously improve the performance of the fusion task and a downstream detection application.
%We argue that the improvement of the fusion results itself can better prompot the downstream vision tasks.
%Besides, 
%The above-mentioned algorithms are only available for the IVIF task.
%While our FusionBooster only applies general operations on the source image, it can be easily extended to other fusion tasks.
%In contrast, our booster, as a general paradigm, is applicable to more fusion tasks.

\section{The Approach}\label{sec2}
In this section, we introduce the proposed FusionBooster (FB) architecture in detail.
We assume that the source images for an arbitrary fusion method at stage one are $I_{\textrm{A}}$ and $I_{\textrm{B}}$.
For example, in the MEIF task, the $I_{\textrm{A}}$ and $I_{\textrm{B}}$ correspond to the underexposed and overexposed images, respectively.
For the backbone method, its initial fusion result at stage one is denoted as $F_{\textrm{init}}$.

\subsection{Problem Formulation}
\label{sectionProblemFormulation}
In the image fusion field, different fusion tasks pursue the same objective, which is to preserve information from different modalities or images with different capture settings.
%In previous works, the formulation of the fusion problem and the core objective of fusion are not fully aligned. 
According to this objective, in our approach, we use the information probe to control the fusion process so as to enhance the relevant information from the source images and thus boost the performance.
%and intensify the image fusion study.

As shown in Fig.~\ref{figureOverallPipeline}, our FB follows the divide and conquer strategy, \ie, the different components of the fusion result are first separated and then enhanced.
Specifically, in the training phase, the information probe learns to gauge the information conveyed by the source images from the initial result outputted by the backbone, which is formulated as:
\begin{equation}
\label{eqProbe}
    [I_\textrm{partA},I_\textrm{partB}] = PU(F_{\textrm{init}}),
\end{equation}
%\textcolor{red}{OK. I now see what you are doing, but it is not explained in a clear way. There is a confusion between the probe and the booster. You are proposing a network which takes the output of the backbone, decomposes it so that you can measure distortion of the source images by the backbone, and then reassembles the decomposed images to produce a new fused image. In the forward pass the network works as a probe, which measures the deviation from I A, I B and F ini. In the backward pass it acts as a booster by adjusting the parameters of the network. Please get rid of eq. 3 and 4. They do not help. You also refer to F bar. It is not defined and you do not need it. Please make it explicit that the losses help to optimise the parameter of branches A, B and ASE Please edit the section accordingly. }
where $PU$ indicates the probe unit, and  $I_\textrm{partA}$ and $I_\textrm{partB}$ represent the underexposed and overexposed components, respectively.
With the probe information in hand, the ASE module is tasked to optimize the assembly of the extracted components to rebuild the initial fusion result $F_{\textrm{init}}$, \textit{i.e.}
\begin{equation}
\label{eqAssembly}
    \hat{F}_{\textrm{init}} = ASE(I_\textrm{partA},I_\textrm{partB}),
\end{equation}
where $\hat{F}_{\textrm{init}}$ denotes the assembly result.

\begin{figure*}[t]
\begin{center}
\includegraphics[width=0.8\linewidth]{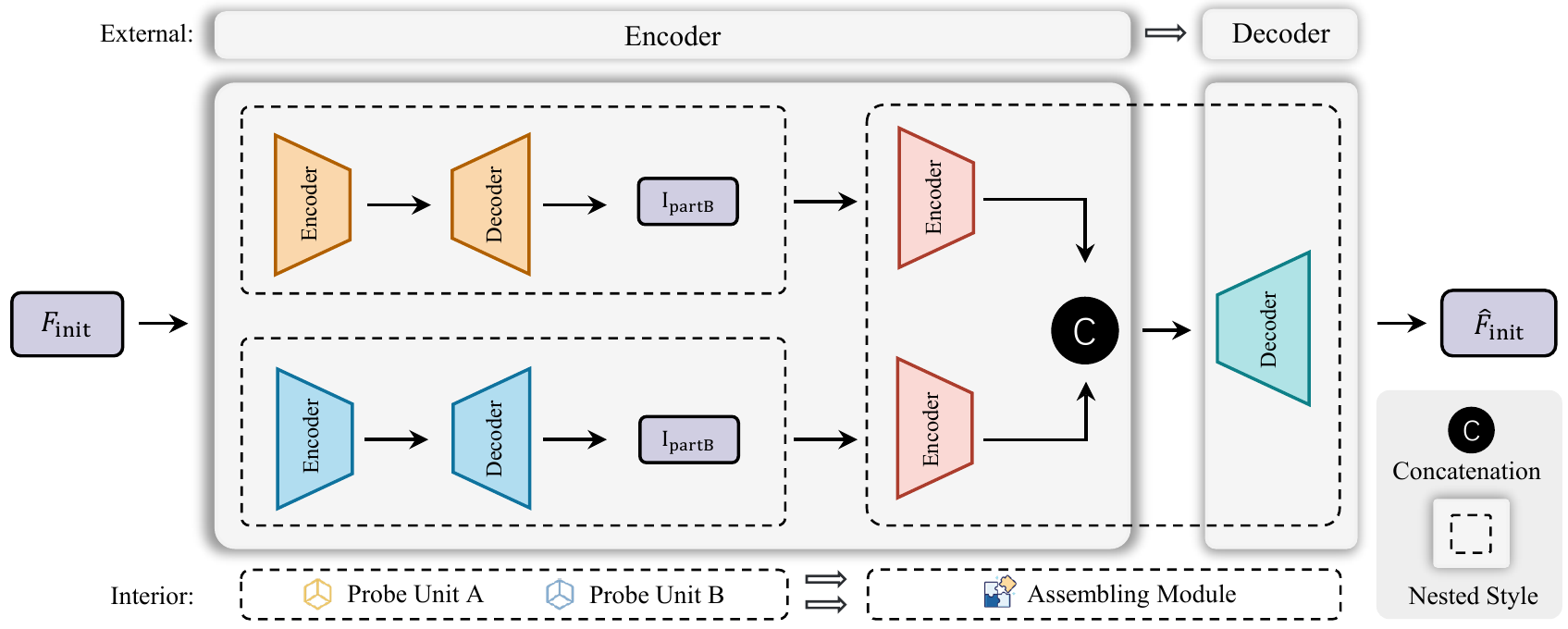}
\end{center}
   \caption{The architecture of the proposed nested AE network.
   The core of this model is composed of the probe units and the assembling module.
   We use the AE-based architecture to formulate these components.
   On the other hand, from an overall (external) perspective, our model can be regarded as an AE to reconstruct the initial fusion result $F_{\mathrm{init}}$.
   The encoder and decoder modules of our network consist of several convolutional layers.
   %The parameters for different encoders and decoders are independent, which will be trained separately.
   %$F_{\textrm{init}}$ is the initial fusion result generated by the backbone method at stage one and it will be utilized to generate the source components $I_{\textrm{partA}}$ and $I_{\textrm{partB}}$.
   %The ProNets are used to perceive the information of source images from the initial result $F_{\textrm{init}}$.
   %While the ASE module controls the integration of the detached components and produce the reconstructed result $\hat{F}_{\textrm{ini}}$.
   %Numbers next to the convolution layers denote the amount of channels of the output feature maps. 
   }
\label{figureNetworkArchitecture}
\end{figure*}

\begin{figure}[t]
\begin{center}
\includegraphics[width=1\linewidth]{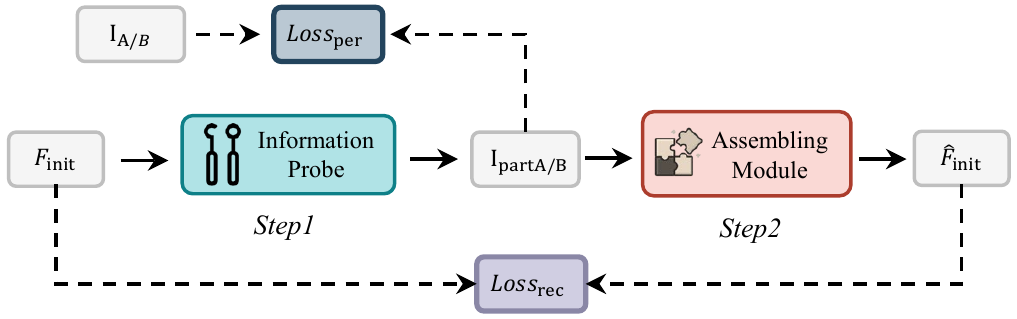}
\end{center}
   \caption{An illustration of the training process of the FusionBooster.   
   The information probe learns to perceive the source images by utilizing a perception loss function.
   The ASE module optimizes the reconstruction loss to rebuild the initial fusion result.
   %We alternately train these two parts to obtain 
   %The information probe first learns to 
   }
\label{traingPhase}
\end{figure}

%We use two perception loss functions 
Given an ideal fusion result, the detached parts in Eq.~\eqref{eqProbe} are expected to obey the following constraints:
\begin{equation}
    I_\textrm{partA} = I_\textrm{A},\ I_\textrm{partB} = I_\textrm{B}.
\end{equation}
However, the information loss issues and the artifacts contained in $F_{\textrm{init}}$ will contaminate these parts and make them degraded.
Thus, in the test phase, we devise a booster layer to recover these two defective components and improve the assembly result.
%and generate the enhanced components $\hat{I}_\textrm{partA}$ and $\hat{I}_\textrm{partB}$. 
%Finally, these refined parts will be further assembled to yield the final output, which can be formulated as:
%where $\hat{F}_{\textrm{ini}}$ denotes the enhanced fused image, $ASE$ is the assembling module.
Since we expect the $\hat{F}_{\textrm{init}}$ to approximately contain all the information from the source images (approach the ideal fused image),  we set to achieve this objective in the booster layer by maintaining the upgraded components and source images as close as possible, \textit{i.e.}
\begin{equation}
    \hat{I}_\textrm{partA} \approx I_\textrm{A},\ \hat{I}_\textrm{partB} \approx I_\textrm{B},
\end{equation}
where $\hat{I}_\textrm{partA}$ and $\hat{I}_\textrm{partB}$ indicate the boosted components.
In this way, the enhanced $F_{\textrm{init}}$ will become more informative and have refined imaging quality.

Without considering the weight measurement of source images, we only focus on strengthening the perceived parts of the initial result.
Thus, compared to the conventional approach with one stage being used to handle multiple issues, our divide and conquer strategy has distinct advantages.

\subsection{ FusionBooster training}
%In the training process, our booster involves the optimization of the information probe and the ASE module.
The trainable parameters of our FB are from the information probe and the ASE module.
Essentially, our FB only involves reconstruction tasks in the training process.
%\textbf{Network architecture:} 
%Our networks in the information probe are used to perceive the source components from the $F_{\textrm{init}}$.
%While the ASE module will assemble these components together to , we reproduce the $F_{\textrm{init}}$ in .
Thus, as we discussed in Section~\ref{related_work_learning}, we use the autoencoder(AE) architecture to implement the ASE module and the probe units.
As shown in Fig.~\ref{figureNetworkArchitecture}, our network can be regarded as a nested AE network.
Specifically, from the external point of view, our FusionBooster architecture is reconstructing the initial result by using the  information probe.
From the internal view, the information probe and the ASE module are using three AE networks to divide and enhance the initial fused image.
Here, the encoder and decoder parts of this network are composed of several convolutional layers.

In Fig.~\ref{traingPhase}, we present the iterative training paradigm of our FusionBooster.
Specifically, we use two loss functions
%, \textit{i.e.},  $Loss_{\textrm{perA}}$, $Loss_{\textrm{perB}}$, and $Loss_{\textrm{rec}}$ 
to perceive the source components and reconstruct the initial fusion result at the pixel level.
Thus, the total loss can be defined as:
\begin{equation}
\label{loss_tot}
Loss_{\textrm{total}} = Loss_{\textrm{per}} + Loss_{\textrm{rec}},
\end{equation}
where  $Loss_{\textrm{per}}$ and $Loss_{\textrm{rec}}$ indicate the perception loss and the reconstruction loss. 

In the information probe, since we have to handle the diversity of the source images among different fusion tasks, we assume the perceived images are of equal importance.
Accordingly, the two probe units use the identical network structure, but their parameters are not shared.
The corresponding loss function is formulated as:
\begin{equation}
Loss_{\textrm{per}} = Loss_{\textrm{perA}}+Loss_{\textrm{perB}},
\end{equation}
\begin{equation}
Loss_{\textrm{perA}} = \frac{1}{H W}\sum\limits_{i}\sum\limits_{j} \lvert I_{\textrm{partA}}(i,j)-I_{A}(i,j) \rvert,
\end{equation}
\begin{equation}
Loss_{\textrm{perB}} = \frac{1}{H W}\sum\limits_{i}\sum\limits_{j} \lvert I_{\textrm{partB}}(i,j)-I_{B}(i,j) \rvert,
\end{equation}
where $H$ and $W$ denote the height and width of the images.

On the other hand, the ASE module is responsible for piecing these detached components together to deliver the reconstructed initial result.
%in the ASE module, we extract the features of the source components and integrate them in the encoder part, and the subsequent convolution layers in the decoder are used to produce the updated fusion result.
%All the convolution layers use the standard $3\times 3$ kernels with the stride of 1.
%Besides, we use the reflection padding operation to keep the resolution of the feature maps constant.
%In this way, our FB accepts fusion results with arbitrary image size.
We train it in the second step, keeping all the parameters in the information probe frozen.
The corresponding reconstruction loss function used to optimize this module is defined as:

\begin{equation}
Loss_{\textrm{rec}} = \frac{1}{H W}\sum\limits_{i}\sum\limits_{j} \lvert \hat{F}_{\textrm{init}}(i,j)-F_{\textrm{init}}(i,j) \rvert.
\end{equation}
Since we do not apply complicated transformations or constrain the detached components in the feature domain by using the pre-trained model~\citep{long2021rxdnfuse,xu2020u2fusion}, the ASE module can smoothly rebuild the initial result and extra computational burden can be avoided.

\begin{figure}[t]
\begin{center}
\includegraphics[width=1\linewidth]{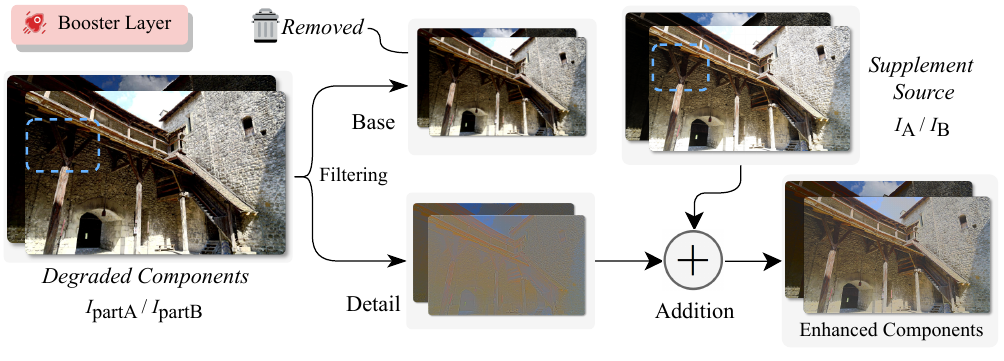}
\end{center}
   \caption{An illustration of the booster layer.
   As shown in the highlighted regions, the decomposed components are unable to recover the information from the source images perfectly.
   Based on the supplementary source images and the image sharpening technique, this layer is designed to enhance these degraded constituents.
   %Thus, we decompose them by using the filtering technique. The base layer is simply discarded and replaced with the high-quality information from the supplement source.
   %Meanwhile, we use the high frequency information to indicate the fusion styles of backbone method and mix the supplement source and the detail layers together to yield the enhanced components.%$\hat{I}_{\textrm{partA}}$ and $\hat{I}_{\textrm{partB}}$.
   %The enhanced images is able to contain more details in the dark regions of the house and the illumination is also improved.
   }
   \vspace{-1mm}
\label{figureBoostLayer}
\end{figure}

\subsection{Booster Layer}
\label{SecBoosterLayer}
%In our FB, the basic idea is to mitigate the information loss issue during the fusion process of the first stage.
%Once the information probe and the ASE module are well trained, 
The booster layer is designed to improve the quality of the fused image.
Simultaneously, it preserves the fusion style of the backbone method, which is embedded within the detached components.
%The booster layer is designed to improve the quality of the fused image while simultaneously preserve the fusion style of the backbone method (hidden in the detached components).
%As shown in Fig.~\ref{figureBoostLayer}, we follow the traditional operation of image sharpening to realize this purpose.
Since we need to cover multiple fusion tasks,  the flexibility would be sacrificed if extra measurements or parameters were introduced in this layer.
Besides, as discussed in Section~\ref{sectionProblemFormulation}, the refined constituent components should approach the source images.
Thus, as shown in Fig.~\ref{figureBoostLayer}, we only use the clean source images $I_{\textrm{A}}$ and $I_{\textrm{B}}$ of different fusion tasks in this layer as the reference sources.
Specifically, for a degraded image component, $e.g.$, $I_{\textrm{partA}}$, we apply average filtering to obtain the low frequency component $I_{\textrm{partA}}^b$ (base layer) as
\begin{equation}
\label{equBaseLayer}
I_{\textrm{partA}}^b = I_{\textrm{partA}}*D(k),
\end{equation}
where $D(k)$ denotes the average filter with the size of $(2k+1)\times (2k+1)$.
Correspondingly, the high-frequency component (the details layer) can be represented as
\begin{equation}
I_{\textrm{partA}}^d = I_{\textrm{partA}} - I_{\textrm{partA}}^b.
\end{equation}

The proposed booster layer is expected to take care of the degraded components.
However, we also need to keep the fusion styles or clues in the output components for the reassembly in the ASE module.
%the as the assembling module is trained based on the initial result, the output images should also preserve the  of the backbone methods.
%so that the assembling module can successfully rebuild the enhanced result.
Thus, we follow the image sharpening operation by combining the clean source image with the detail layer of the degraded component, \textit{i.e.}
\begin{equation}
\hat{I}_{\textrm{partA}} = I_{\textrm{A}} + I_{\textrm{partA}}^d.
\end{equation}
Here, the high-frequency information from the degraded component is expected to provide fusion clues and edge sharpening for the ASE module.
Such enhancement to the edge information has been demonstrated to be useful for the downstream tasks~\cite{cheng2023mufusion,liu2022target}.
Involving the source images in the enhanced component $\hat{I}_{\textrm{partA}}$ helps to replace the degraded base layer with the informative one and forces the ASE module to deliver a more robust fusion result.
The effectiveness of the booster layer design will be demonstrated in Section~\ref{sectionAblation}.
%and the combination of these two items will generate the enhanced source input for the proxy mode to deliver high-quality fusion result.
%Therefore, in the booster layer, we follow two principles while enhancing the input of the agent model, \textit{i.e.}, the preservation of the.
%while the output of final result will be correspondingly improved.

\section{Experiment}
\subsection{Experimental Settings}
We apply our FB to three widely investigated image fusion tasks, \textit{i.e.}, the IVIF task, the MFIF task, and the MEIF task.
%We apply our FB to three widely investigated image fusion tasks, \textit{i.e.}, IVIF, MFIF, and MEIF.
Three public benchmark datasets are used in our experiments, including the LLVIP dataset~\citep{jia2021llvip} for the IVIF task, MFI-WHU dataset~\citep{zhang2021mffgan} for the MFIF task, and SCIE dataset~\citep{Cai2018medataset} for the MEIF task.

The LLVIP dataset is very challenging.
It is mostly composed of high-quality infrared and visible image pairs in the low-light environment.
The MFI-WHU dataset contains 120 far-focused and near-focused image pairs of different scenes.
The SCIE dataset consists of 590 high-resolution indoor and outdoor image sequences with different exposure settings.
Considering the small scale of the last two datasets, we randomly crop $128\times 128$ patches for augmenting the training data. 
The number of images or patches used for training is 12,025, 33,703, and 11,702, respectively.
The number of randomly selected image pairs used for evaluation is 250, 20, and 51, respectively.
%and we carefully select the registered underexposed and overexposed image pairs from this dataset for the training and evaluation.
%In our experiments, the entire 12,025 image pairs in the training set is used to optimize the models and 250 randomly selected images from the test set is used for evaluation.
%To reduce the time consumptioon of the image fusion methods during the experimetns.
%On the other hand, the MFI-WHU dataset contains 120 far-focused and near-focused image pairs in different scenes.

\begin{figure}[t]
\begin{center}
\includegraphics[width=1\linewidth]{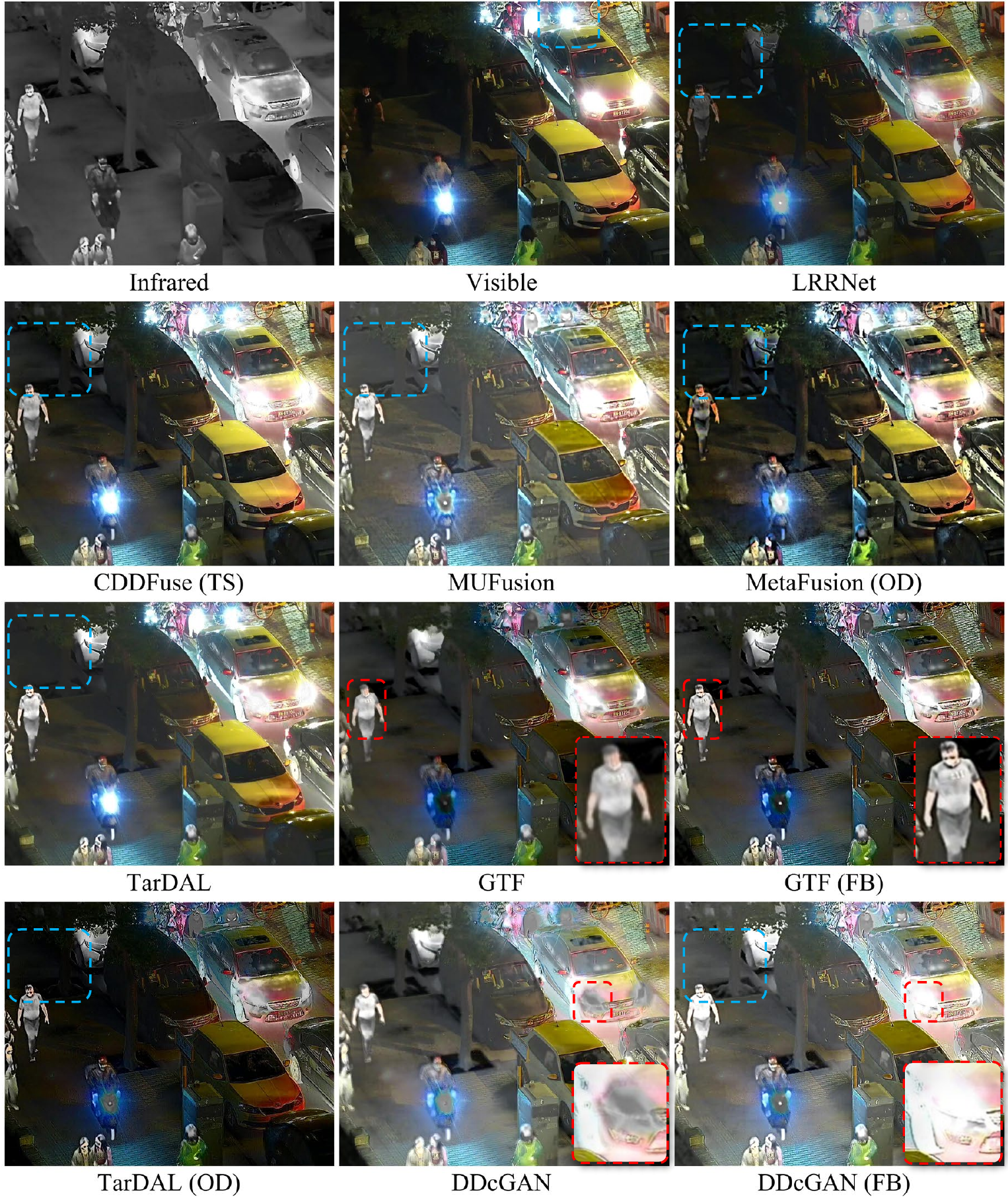}
\end{center}
   \caption{Illustration of the qualitative results of the infrared and visible image fusion  on one pair of images from the LLVIP dataset.}
   %The GTF and DDcGAN are traditional method and learning based method, respectively.
   %As denoted by the yellow arrows, fusion results of GTF will have better visualization results on different areas, $e.g.$, the pedestrians and the bike.
   %For the learning-based method DDcGAN, its fusion results contain some artifacts around the target areas.
   %Our FB can address this issue by presenting results with higher image quality.
   %(FB: FusionBooster; Zoom in for better view.)}
\label{LLVIP_qualitative_pedestrian}
\end{figure}

%After that, the number of patches used to train the MF and ME models is 33,703 and 11,702 respectively.
%During the test phase, the number of images used in these two tasks are 20 and 51 respectively and the whole images are fed into the FusionBooster to yield the enhanced output.

This algorithm is implemented in PyTorch and executed on  an NVIDIA GeForce RTX 3090 GPU.
The Adam optimizer~\citep{kingma2014adamOptimizer} is used to update the parameters of the models with the learning rate of $10^{-4}$.
The number of epochs is set as 10 and the batch size is 2.
The filter size $k$ in Eq.~\eqref{equBaseLayer} is empirically set as 3.
All the competitors' implementations come from the code repositories mentioned in the original papers or reproduced by other researchers.
%\textit{More experimental results can be found in the supplementary materials.}

For the quantitative experiments, five widely used image fusion metrics, \textit{i.e.}, visual information fidelity (VIF)~\citep{han2013viff}, an objective image fusion performance measure ($Q_{abf}$)~\citep{xydeas2000qabfAndEI}, information entropy (EN)~\citep{roberts2008metricEN}, edge intensity (EI)~\citep{xydeas2000qabfAndEI}, and standard deviation (SD)~\citep{cheng2021unifusion} are adopted to evaluate the fusion performance from different perspectives.
Specifically, VIF measures the distortion between the fusion result and the source images to indicate the information fidelity.
$Q_{abf}$ is used to measure the preservation ability of the gradient information from the input images.
%Higher values of EN and SD indicate that the image has better performance in the perspectives of information amount and contrast.
EN and SD measure the information content and contrast of the image. 
%that the image has better performance in the perspectives of 
Finally, the edge information and clarity of the fusion results are reflected by EI.
%The MFI-WHU dataset contains 120 far-focused and near-foucsed image pairs.
%To obtain more data for the network training, we randomly select 100 image pairs from this dataset 

%Both the infrared and visible image fusion task can be used in this lowlight environment, thus we use this...

\begin{figure}[t]
\begin{center}
\includegraphics[width=1\linewidth]{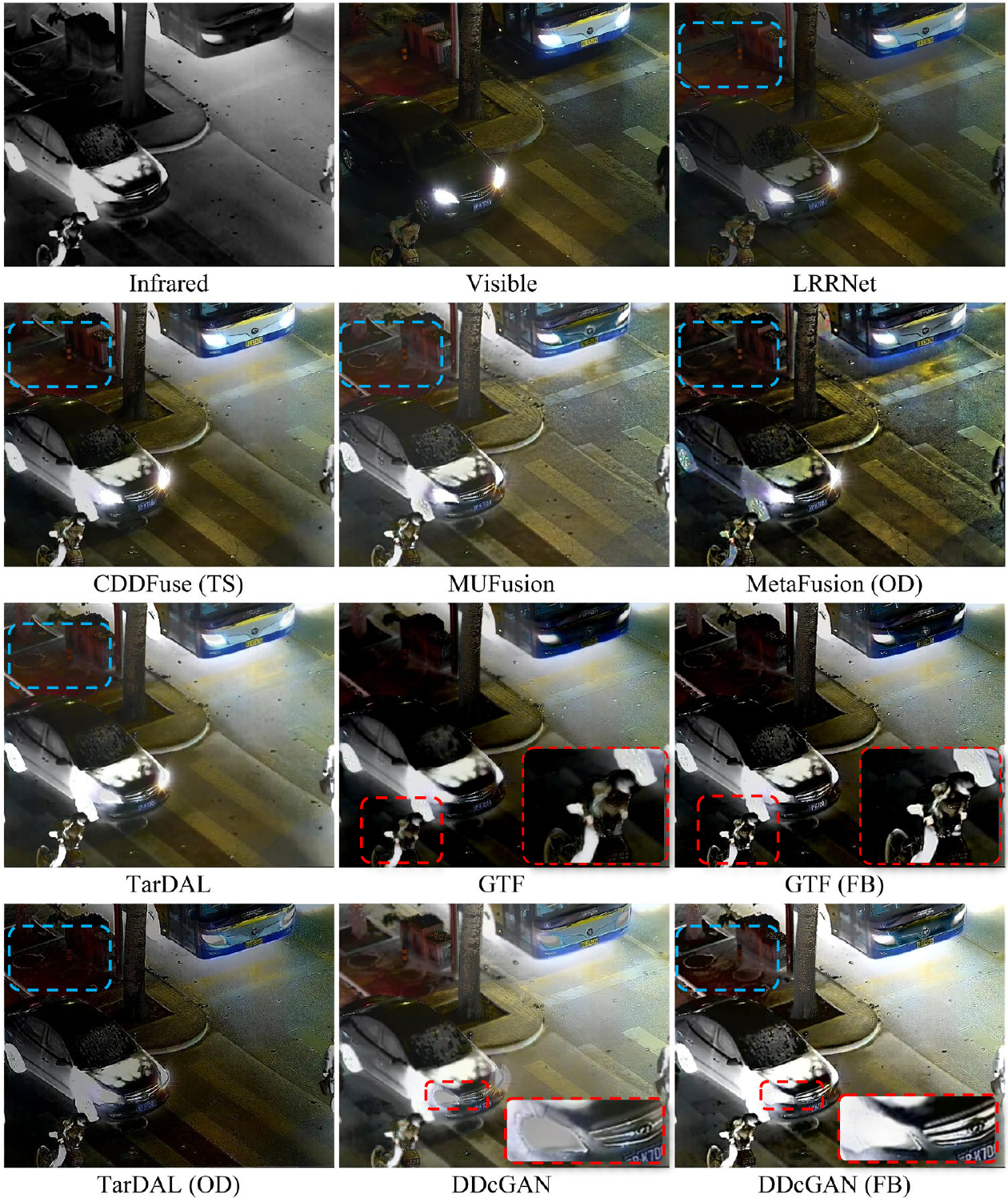}
\end{center}
   \caption{Illustration of the qualitative results of the infrared and visible image fusion on another pair of images from the LLVIP dataset.}
   %The GTF and DDcGAN are traditional method and learning based method, respectively.
   %As denoted by the yellow arrows, fusion results of GTF will have better visualization results on different areas, $e.g.$, the pedestrians and the bike.
   %For the learning-based method DDcGAN, its fusion results contain some artifacts around the target areas.
   %Our FB can address this issue by presenting results with higher image quality.
   %(FB: FusionBooster; Zoom in for better view.)}
\label{LLVIP_qualitative_bike}
\end{figure}

% Table generated by Excel2LaTeX from sheet 'w_o_PSNR_SwinFusion (2)'

% Table generated by Excel2LaTeX from sheet 'w_o_PSNR_SwinFusion (2)'
\begin{table*}[tbp]
  \centering
  \caption{The quantitative results obtained by the proposed FusionBooster on the LLVIP dataset, compared with other methods W/O extra model or stage. (\textbf{Bold}: Best; \underline{Underline}: Second best)}
    \label{table_quantativeLLVIP_nonEnhancement}
  \resizebox{0.6\linewidth}{!}{ 
    \begin{tabular}{ccccccc}
    \toprule
    Method & Venue & SD    & EN    & VIF   & EI    & Qabf \\
    \midrule
    GTF   & 16' Inf. Fus. & 50.164  & 7.351  & 0.576  & 44.129  & 0.454  \\
    U2Fusion & 20' TPAMI & 37.428  & 6.707  & 0.492  & 46.899  & \underline{0.499}  \\
    DDcGAN & 20' TIP & 51.495  & \underline{7.431}  & 0.764  & 49.127  & 0.395  \\
    SD-Net & 21' IJCV & 36.257  & 6.889  & 0.414  & 44.609  & 0.482  \\
    ReCoNet & 22' ECCV & 48.761  & 5.962  & 0.727  & 47.178  & 0.462  \\
    TarDAL & 22' CVPR & \underline{52.106}  & 7.353  & \underline{0.809}  & 46.126  & 0.444  \\
    YDTR  & 22' TMM & 36.502  & 6.782  & 0.388  & 31.336  & 0.334  \\
    LRRNet & 23' TPAMI & 29.826  & 6.423  & 0.342  & 34.928  & 0.406  \\
    MUFusion & 23' Inf. Fus. & 40.104  & 7.019  & 0.755  & \underline{57.698}  & \textbf{0.547 } \\
    \rowcolor[rgb]{ .851,  .851,  .851} YDTR (FB) & Ours  & 41.159  & 6.988  & 0.532  & 48.590  & 0.470  \\
    \rowcolor[rgb]{ .851,  .851,  .851} GTF (FB) & Ours  & {53.260}  & 7.412  & {0.884}  & {73.185}  & 0.485  \\
    \rowcolor[rgb]{ .851,  .851,  .851} DDcGAN (FB) & Ours  & \textbf{57.672 } & \textbf{7.650 } & \textbf{0.986 } & \textbf{75.362 } & 0.470  \\
    \bottomrule
    \end{tabular}%
}
\end{table*}%

% Table generated by Excel2LaTeX from sheet 'w_o_PSNR_SwinFusion (2)'
\begin{table*}[tbp]
  \centering
  \caption{The quantitative results obtained by the proposed FusionBooster on the LLVIP dataset, compared with other methods W/ extra model or stage. (TS: Two-stage; OD: Object detection; Seg: Segmentation; LE: Low-light enhancement; FB: FusionBooster)}
    \label{table_quantativeLLVIP_Enhancement}
  \resizebox{0.75\linewidth}{!}{  
    \begin{tabular}{cccccccc}
    \toprule
    Method & Venue & Model/Stage (MB) & SD    & EN    & VIF   & EI    & Qabf \\
    \midrule
    RFN-Nest & 21' Inf. Fus. & TS (17.179) & 39.719  & 7.064  & 0.466  & 34.195  & 0.384  \\
    TarDAL++ & 22' CVPR & OD (14.46) & 41.059  & 6.604  & 0.676  & 70.005  & 0.367  \\
    SeAFusion & 22' Inf. Fus & Seg (\underline{0.646}) & 51.810  & 7.451  & 0.839  & 55.935  & \underline{0.618}  \\
    DIVFusion & 23' Inf. Fus & LE (225.668) & \underline{53.370}  & \underline{7.556}  & \underline{1.234}  & 56.595  & 0.349  \\
    CDDFuse & 23' CVPR & TS (1.462) & 50.409  & 7.374  & 0.787  & 52.324  & \textbf{0.622 } \\
    MetaFusion & 23' CVPR & OD (14.46) & 49.935  & 7.148  & \textbf{1.539 } & \textbf{81.840 } & 0.436  \\
    \rowcolor[rgb]{ .851,  .851,  .851} DDcGAN (FB) & Ours  & FB (\textbf{0.168}) & \textbf{57.672 } & \textbf{7.650 } & 0.986  & \underline{75.362}  & 0.470  \\
    \bottomrule
    \end{tabular}%
}
\end{table*}%

\subsection{An Infrared and Visible Image Fusion Task}
In this section, we present the fusion results obtained by advanced image fusion methods and some algorithms enhanced by our booster.
As it is an important task in the image fusion field, we select more competitor algorithms for the comparative evaluation.
The tested algorithms include the traditional method GTF~\citep{ma2016gtf}, 5 CNN-based methods, namely  U2Fusion~\citep{xu2020u2fusion}, SDNet~\citep{zhang2021sdnet}, ReCoNet~\cite{huang2022reconet}, LRRNet~\cite{li2023lrrnet} and MUFusion~\citep{cheng2023mufusion}, 6 approaches that contain additioanl enhancement model or fusion stage, $i.e.$, RFN-Nest~\cite{li2021rfn}, TarDAL++~\cite{liu2022target}, SeAFusion~\citep{tang2022SeAFusion}, DIVFusion~\cite{tang2023divfusion}, CDDFuse~\cite{zhao2023cddfuse}, and MetaFusion~\cite{zhao2023metafusion}, the GAN-based method DDcGAN~\citep{ma2020ddcgan}, and the transformer-based method YDTR~\citep{tang2022ydtr}.

%\subsubsection{Qualitative experiments}
\subsubsection{Qualitative Experiments}
For the IVIF task, due to the limitations of the handcrafted image features, the traditional methods cannot handle complex scenes effectively.
As shown in Fig.~\ref{LLVIP_qualitative_pedestrian} and Fig.~\ref{LLVIP_qualitative_bike}, the traditional method, GTF, suffers from the blurring issues in the fusion results.
Our booster can effectively address this and produce visually pleasing images.
Meanwhile, our paradigm also reduces the artifacts, which severely degrade the image quality of DDcGAN.
%Besides, the characteristics of the original fusion methods are well preserved, \textit{e.g.}, the salient thermal information.
Besides, compared with the SOTA methods LRRNet and TarDAL, the enhanced DDcGAN inherits the merits of the original method and shows the ability to cope with the challenges of dark environments, preserving the details of the background (blue boxes), and presenting more salient thermal information on the foregrounds. 
Finally, when the object detection model is used to enhance the TarDAL, compared with the original method, the fusion results of this method show a lack of brightness in the background and the thermal radiation in the target regions, which is consistent with our discussion about the task gap issue.
In Section~\ref{more_IVIF}, we further demonstrate the impact of our booster on TarDAL. The results indicate that our approach is able to mitigate this issue.

\begin{figure}[t]
\begin{center}
\includegraphics[width=1\linewidth]{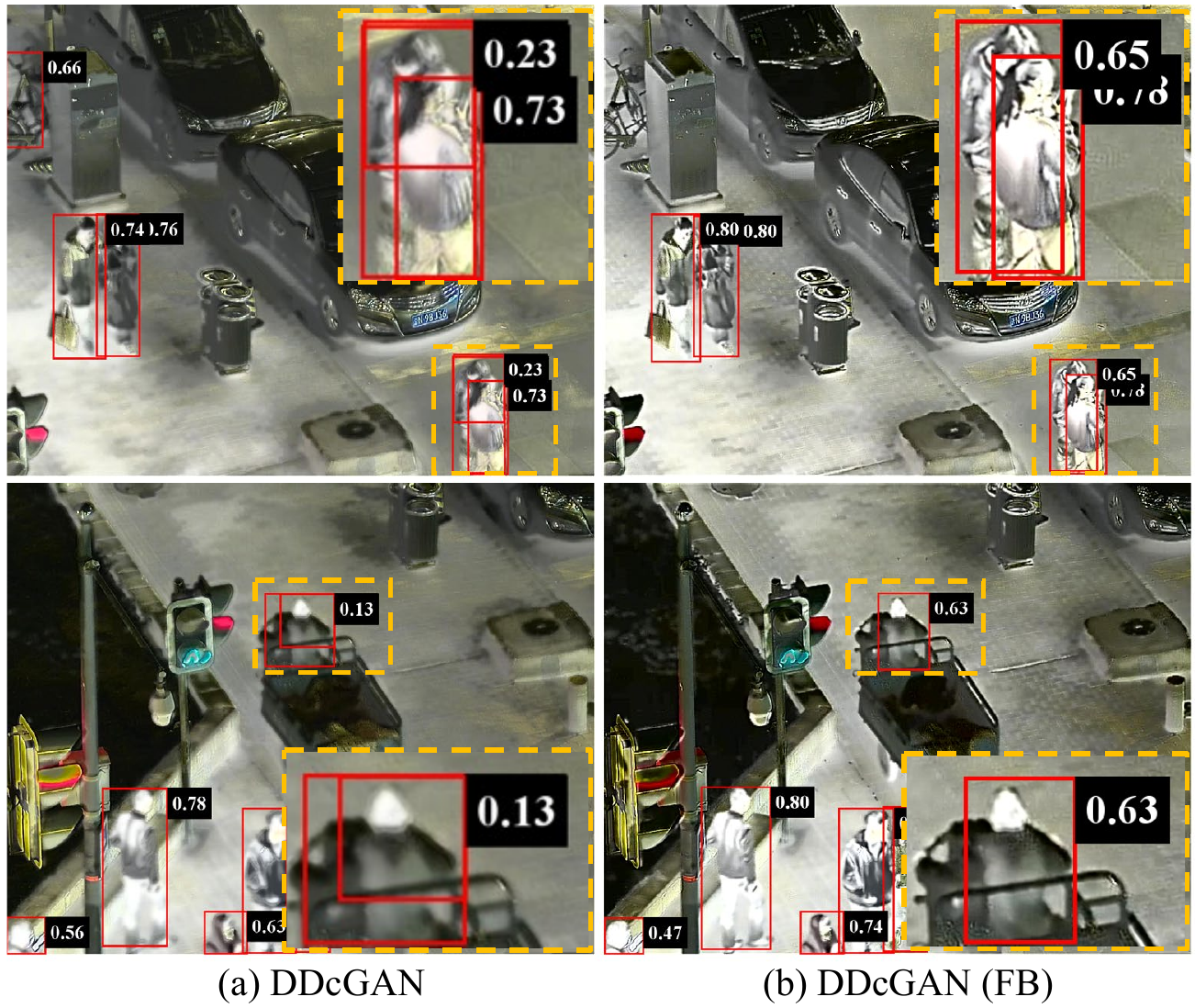}
\end{center}
   \caption{Visualization of the results obtained by DDcGAN and DDcGAN with FusionBooster on the pedestrian detection task.}
\label{LLVIP_detection_qualitative}
\end{figure}

%\subsubsection{Quantitative experiments}
\subsubsection{Quantitative Experiments}
For the quantitative comparison, we select three different types of fusion methods, \textit{i.e.}, the traditional method GTF, the transformer-based method YDTR, and the GAN-based method DDcGAN as the backbone methods of our booster.
Meanwhile, for the competitors, we also divide them into two categories, \ie, methods with or without using an extra model or stage.
As shown in Table~\ref{table_quantativeLLVIP_nonEnhancement}, our booster consistently improves the performance of various types of algorithms on all of these five metrics.
The remarkable performance on these metrics indicates that the proposed booster is able to increase the fidelity of the information derived from the source images (VIF), better preserve the gradient information (Qabf), and produce robust fused images with sharp edge information (EN, SD, and EI).
Moreover, the DDcGAN, proposed in 2020 and upgraded by our FusionBooster, outperforms current SOTA methods in 4 out of 5 metrics, which is a significant improvement.

In addition, we use the upgraded DDcGAN to conduct further experiments involving other methods, with an extra stage or enhancement module.
As shown in Table~\ref{table_quantativeLLVIP_Enhancement}, our FusionBooster has the smallest volume compared to other enhancement models.
For the quantitative results, the MetaFusion exhibits a similar performance as our upgraded method.
However, as shown in Fig.~\ref{LLVIP_qualitative_pedestrian} and Fig.~\ref{LLVIP_qualitative_bike}, its inability to address the task gap issue results in poor performance on the metrics of SD and EN.
By contrast, our fusion results effectively achieve a balance between the image quality, and the correlation with the source images, obtaining the best performance on the non-reference metrics and comparable results on the VIF and Qabf.
The best overall performance in this comparison also demonstrates that there is a scope for the existing image fusion methods to achieve performance gains without considering other vision tasks.
%The best overall performance on this comparison demonstrates that the current image fusion performance can be still improved without the consideration of other vision tasks.
%although some advanced algorithms, \textit{e.g.}, TarDAL, are trained jointly with other tasks, the DDcGAN with our booster outperforms both the TarDAL (optimized for the human vision) and TarDAL++ (optimized for both vision and detection) models on all of these five metrics.

%We can observe that, after the combination of the detection and fusion task, the performance of this method is not very promising.

%For the quantitative results, a SOTA method TarDAL is presented in the table.
%As shown in this table, although this method is trained combined with downstream fusion task, its fusion ability cannot match...
%The optimal visualization and detection model is selected as the competitor.

% Table generated by Excel2LaTeX from sheet 'Sheet1 (3)'
\begin{table}[tbp]
%\renewcommand{\arraystretch}{1}
%\footnotesize
  \centering
  \caption{The accuracy of pedestrian detection using different modalities on the LLVIP dataset.}
    \label{table_quantative_pedestrian}
  \resizebox{0.8\linewidth}{!}{  
    \begin{tabular}{cccc}
    \toprule
    Method & Venue & $mAP_{50:95}$(\%) & $mAP_{50}$(\%) \\
    \midrule
    Visible & Input & 54.2  & 94.4  \\
    DDcGAN & 20' TIP & 63.7  & 94.4  \\
    DIVFusion (LE) & 23' Inf. Fus. & 64.2  & 97.1  \\
    YDTR  & 22' TMM & 64.9  & 97.4  \\
    CDDFuse (TS) & 23' CVPR & 65.4  & 97.0  \\
    GTF   & 16' Inf. Fus & 65.7  & 96.5  \\
    MetaFusion (OD) & 23' CVPR & 66.1  & 97.3  \\
    MUFusion & 23' Inf. Fus. & 66.4  & 96.2  \\
    DenseFuse & 18' TIP & 66.5  & 96.4  \\
    LRRNet & 23' TPAMI & 66.5  & 97.0  \\
    TarDAL & 22' CVPR & 66.6  & 96.9  \\
    SeAFusion (Seg) & 22' Inf. Fus. & 66.9  & 97.2  \\
    U2Fusion & 20' TPAMI & 67.3  & 97.0  \\
    Infrared & Input & 67.9  & 97.3  \\
    SDNet & 21' IJCV & 68.1  & 97.3  \\
    TarDAL++ (OD) & 22' CVPR & 68.3  & 97.2  \\
    \rowcolor[rgb]{ .851,  .851,  .851} GTF (FB) & Ours  & 67.8 (+2.1) & 97.0 (+0.5) \\
    \rowcolor[rgb]{ .851,  .851,  .851} YDTR (FB) & Ours  & 67.6  (+2.7) & \textbf{97.9 (+0.5)} \\
    \rowcolor[rgb]{ .851,  .851,  .851} DDcGAN (FB) & Ours  & \textbf{69.3 (+5.6)} & 97.4 (+3.0) \\
    \bottomrule
    \end{tabular}}
    %}
\end{table}%

%多聚焦图像融合的可视化
\begin{figure*}[t]
\begin{center}
\includegraphics[width=1\linewidth]{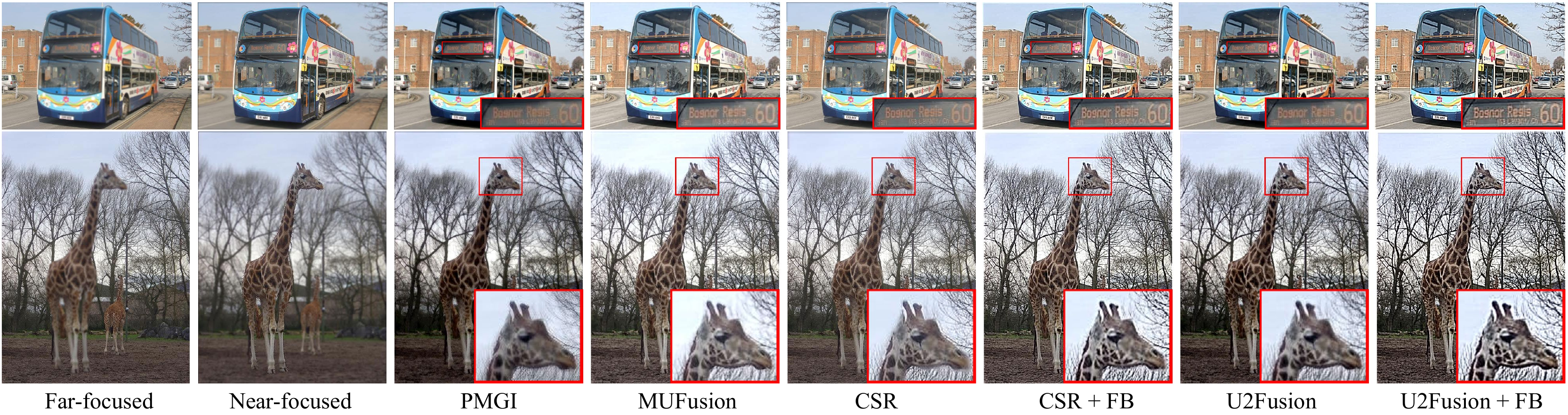}
\end{center}
   \caption{The qualitative results achieved in the MFIF task on two pairs of images from the MFI-WHU dataset.}
   %The selected backbones CSR and U2Fusion are traditional method and learning based method, respectively.
   %As shown in the highlighted regions, both the CSR and U2Fusion have clearer view of the board on the bus.
   %Meanwhile, as denoted by the red and yellow arrows, our FusionBooster successfully increases the clarity of the fusion results on the region of the giraffe' head.}
\label{MFI-WHU_qualitative}
\end{figure*}

\subsubsection{The Pedestrian Detection Task}
%\subsubsection{Pedestrian detection task}
In addition to the visual quality, an important application for the IVIF task is to improve the performance of downstream vision tasks by using the complementary information contained in the fused image.
In this experiment, we use the YOLOv5 detector to test the accuracy of different image fusion methods on the pedestrian detection task.
We separately train the detector by using the fusion results of different algorithms on the training set of the LLVIP dataset.
The trained models are used to detect pedestrians in different modalities.
%Existing image fusion methods only persuade the high quality image fusion results, ignoring the perofrmance of image fusion methods on the downstream fusion task.
As shown in Table~\ref{table_quantative_pedestrian}, in the low-light environment, the accuracy of some SOTA methods cannot even match that of the single modality, $\textit{i.e.}$, the infrared modality.
However, once the FB is applied in conjunction with these methods, the average precision is significantly improved, $\textit{e.g.}$, $5.6\%$ for DDcGAN over the IoU thresholds from 0.5 to 0.95.
%Besides, although the MUFusion has promising results on the image quality assessments, its accuracy on the detection task is not satisfactory.
%This demonstrates that better image quality assessments do not consistently lead to higher performance on downstream tasks.
It is worth noting that the performance of DDcGAN with our booster is better than that of the SeAFusion and TarDAL++, which consider similar segmentation and detection tasks in their training process.

In Fig.~\ref{LLVIP_detection_qualitative}, we present the visualization of two results obtained with our booster in the pedestrian detection task.
The detector has a higher confidence for the detected pedestrians %based on the enhanced fusion results 
and the false detection issues are  mitigated (bike in the top left corner of the first example).
This comparison also reveals that the fusion results with sharpened edge information and higher contrast can benefit the detection task, which is consistent with the motivation of designing the booster layer mentioned in Section~\ref{SecBoosterLayer}.

\begin{figure}[t]
\begin{center}
\includegraphics[width=1\linewidth]{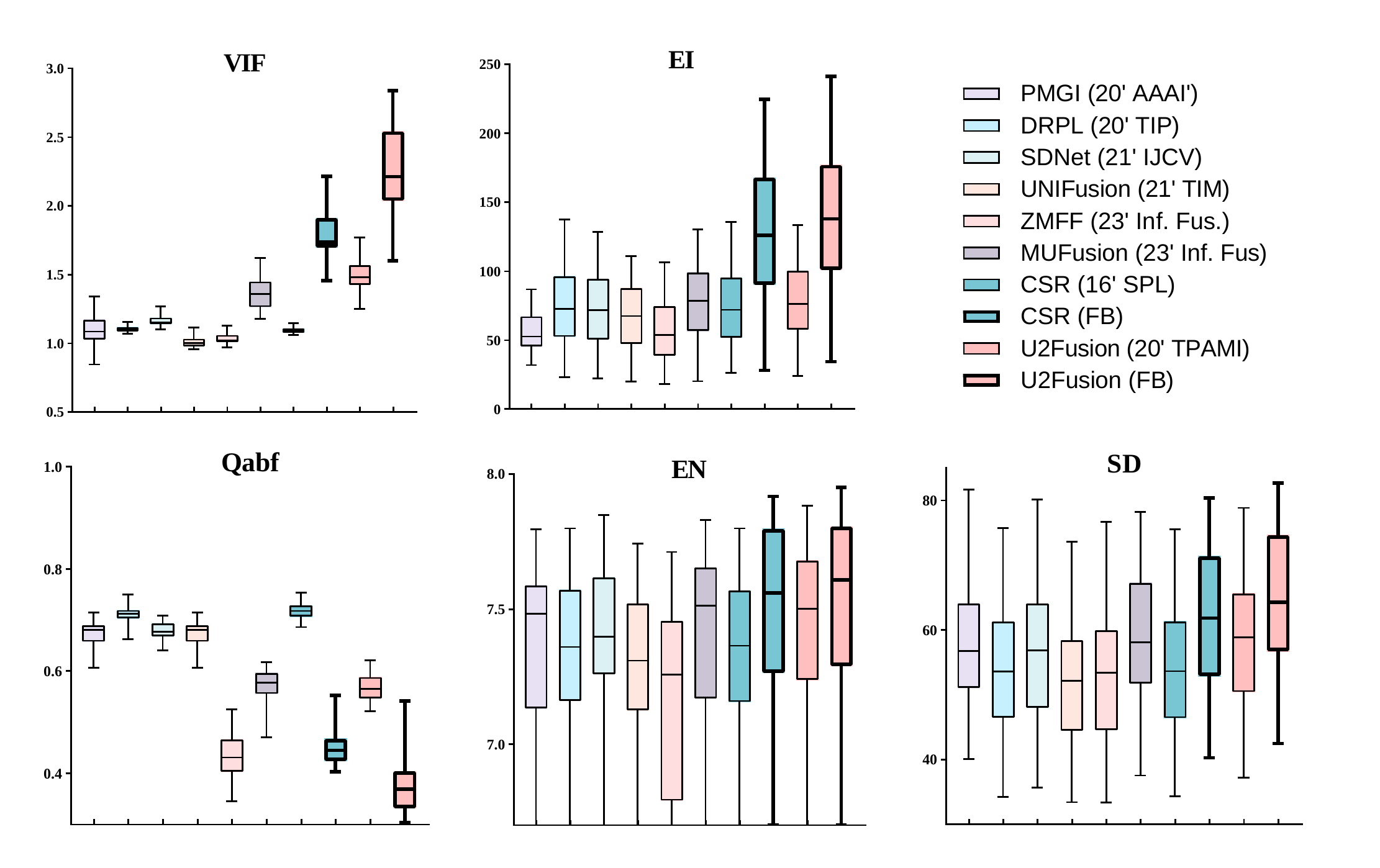}
\end{center}
   \caption{The quantitative results obtained by different fusion methods with (\textbf{bold}) and without the FB in the case of the MFIF task.}
   \vspace{-2mm}
\label{MFI-WHU_quantitative}
\end{figure}

\begin{figure*}[t]
\begin{center}
\includegraphics[width=1\linewidth]{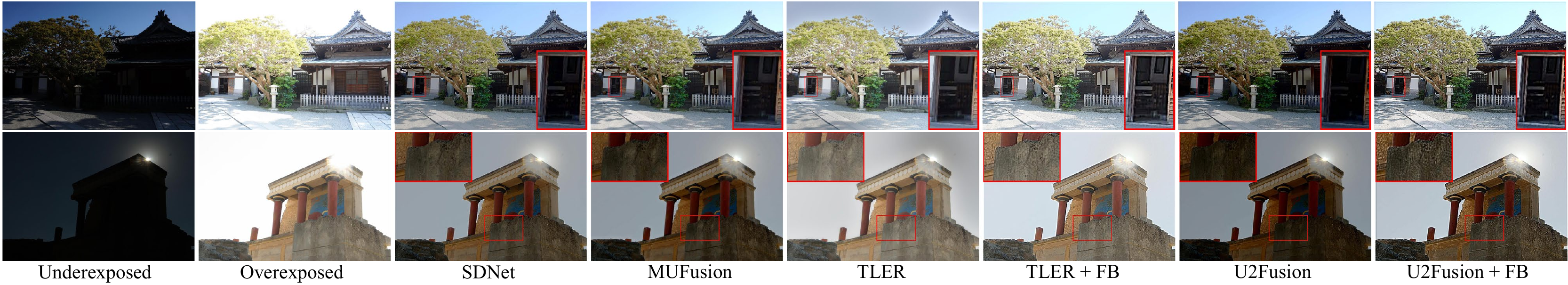}
\end{center}
   \caption{The qualitative results, obtained on two pairs of images from the SCIE dataset, when performing the MEIF task.}
   %The TLER and U2Fusion are traditional method and learning based method, respectively.
   %As denoted by the red arrows, after using the FusionBooster, the clarity of the TLER's fusion results are improved.
   %On the other hand, the fusion results of U2Fusion have more appropriate exposure degree after utilizing the FusionBooster.}
\label{DSCIE_qualitative}
\end{figure*}

\begin{table*}[tbp]
  \centering
  \caption{The quantitative results, obtained on two pairs of images from the SCIE dataset, when performing the MEIF task.}
  \label{table_quantative_MEIF}
  \resizebox{0.6\linewidth}{!}{   
    \begin{tabular}{ccccccc}
    \toprule
    Method & Venue & SD    & EN    & VIF   & EI    & Qabf \\
    \midrule
    DeepFuse & 17' ICCV & 46.091  & 7.155  & 1.295  & 60.526  & 0.696  \\
    TLER  & 18' SPL & 41.443  & 7.103  & 1.713  & 75.570  & \textbf{0.736 } \\
    MEF-GAN & 20' TIP & \underline{52.363 } & 7.192  & 1.592  & 80.214  & 0.373  \\
    U2Fusion & 20' TPAMI & 49.013  & 7.213  & \underline{1.695}  & \underline{83.001}  & 0.638  \\
    SDNet & 21' IJCV & 44.135  & 7.035  & 1.299  & 72.497  & 0.677  \\
    AGAL  & 22' TCSVT & 43.725  & 7.106  & 1.314  & 71.954  & 0.655  \\
    MUFusion & 23' Inf. Fus. & 49.682  & \underline{7.231}  & 1.637  & 70.179  & \underline{0.716 } \\
    IID-MEF & 23' Inf. Fus. & 40.975  & 7.035  & 1.124  & 59.117  & 0.610  \\
    \rowcolor[rgb]{ .851,  .851,  .851} TLER (FB) & Ours  & 50.187  & 7.249 & 1.934 & 110.503 & 0.518  \\
    \rowcolor[rgb]{ .851,  .851,  .851} U2Fusion (FB) & Ours  & \textbf{58.573 } & \textbf{7.506 } & \textbf{2.506 } & \textbf{134.524 } & 0.425  \\
    \bottomrule
    \end{tabular}%
}
\end{table*}%

\subsection{Multi-focus Image Fusion}
In this section, we present the MFIF results obtained by various image fusion methods and two boosting examples of our booster.
For this task, we select 4 methods, \textit{i.e.}, PMGI~\citep{zhang2020rethinking}, DRPL~\citep{li2020drpl}, UNIFusion~\citep{cheng2021unifusion}, and ZMFF~\cite{hu2023zmff} as the competitors.

%\subsubsection{Qualitative experiments}
\subsubsection{Qualitative Experiments}
Due to the general operations used in the booster layer, our FusionBooster is also able to improve existing multi-focus image fusion methods. 
As shown in the first row of Fig.~\ref{MFI-WHU_qualitative}, applying the proposed booster to the traditional method, CSR, and the learning-based method, U2Fusion, the details on the board of the bus become clearer.
Furthermore, in the second example, the original CSR does not accurately infer the focused regions of the source images (head of the ``giraffe').
As shown in the magnified region, the enhanced result successfully addresses this issue by improving the clarity.
Similar conclusion can be reached by looking at the enhancement achieved by U2Fusion.
%Besides, our FusionBooster also improves the learning-based method, U2Fusion, producing a piece of sharper edge information.
In conclusion, compared with the other algorithms, the enhanced methods are superior in terms of preserving the local details in the highlighted area.

%\subsubsection{Quantitative experiments}
\subsubsection{Quantitative Experiments}
\label{SectionMFIFQuantitative}
For the quantitative results, as shown in Fig.~\ref{MFI-WHU_quantitative}, with our booster (legends with bold borders), U2Fusion has a clear advantage over the other advanced methods in terms of  VIF, EI, EN, and SD.
This promising result demonstrates the superiority of the proposed FusionBooster.
Moreover, integrating with our booster, the traditional method CSR~\citep{liu2016csr} also exhibits distinct strengths on multiple metrics.
However, our FB does not perform well on the metric of Qabf. Similar issues also arise in the context of related work~\cite{li2020mdlatlrr,cheng2023mufusion}.
This can be attributed to the enhancement effect of our FusionBooster, which alters the gradient information transferred from the source images into the fusion results.
Consequently, this gradient based metric cannot consistently reflect the benefits of our booster.
%is also revealed in the experimental results of 

\subsection{Multi-exposure Image Fusion Task}
In this section, we present the MEIF results obtained by different methods and some algorithms upgraded with our booster.
Four open source CNN-based MEIF algorithms, \textit{i.e.}, DeepFuse~\citep{prabhakar2017deepfuse}, MEF-GAN~\citep{xu2020mef}, AGAL~\citep{liu2022AGAL_ME}, and IID-MEF~\citep{zhang2023iidmef}, and a traditional method TLER~\citep{yang2018TLER} are involved in the experiments.

\begin{figure*}[h]
\begin{center}
\includegraphics[width=1\linewidth]{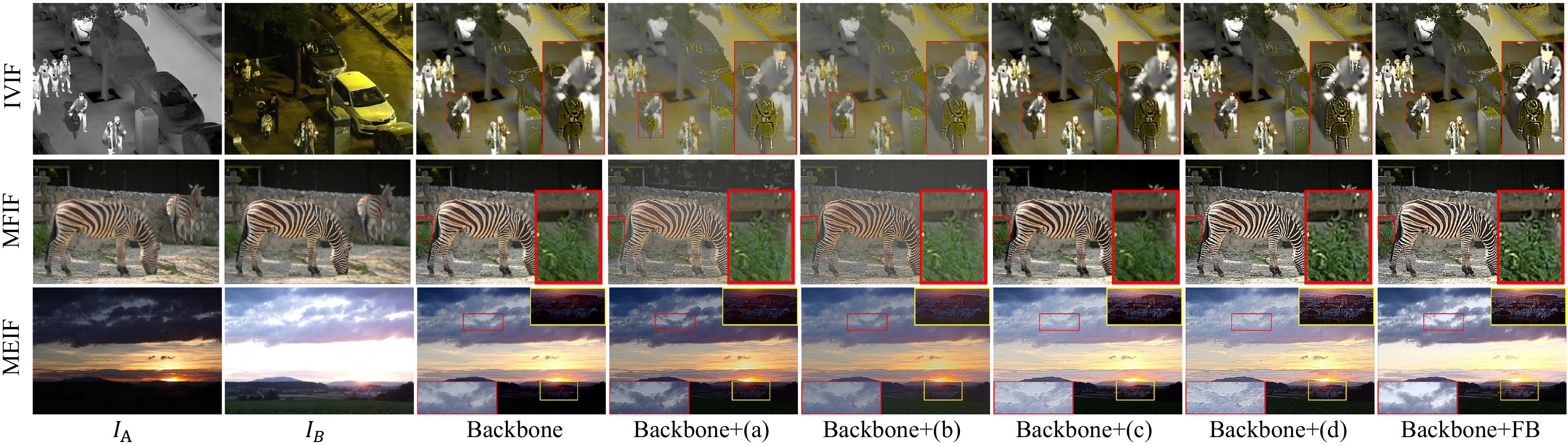}
\end{center}
   \caption{Qualitative results of the ablation experiments.
   Without using the second stage, the explicit enhancement in the first stage (settings (a) and (b)) makes different fusion results blurred in the highlighted regions.
   Meanwhile, without the detached components from the information probe (setting (c) and (d)), the edge information in the IVIF results is not clear and there are some artifacts in the MEIF results.
   The yellow regions in the MEIF results denote the exposure difference in various experimental settings.
   }
   %($I_{\textrm{vis}}$: visible image; $I_{\textrm{ir}}$: infrared image)}
   %The degraded images Since the quality of the reconstructed images is linked to that of the fusion results. The beneficial disturbance contained in the enhanced source images can prompt the rebuilding of the high-quality fused images.}
\label{figureMoreAblationQualitative}
\end{figure*}

\begin{figure}[t]
\begin{center}
\includegraphics[width=1\linewidth]{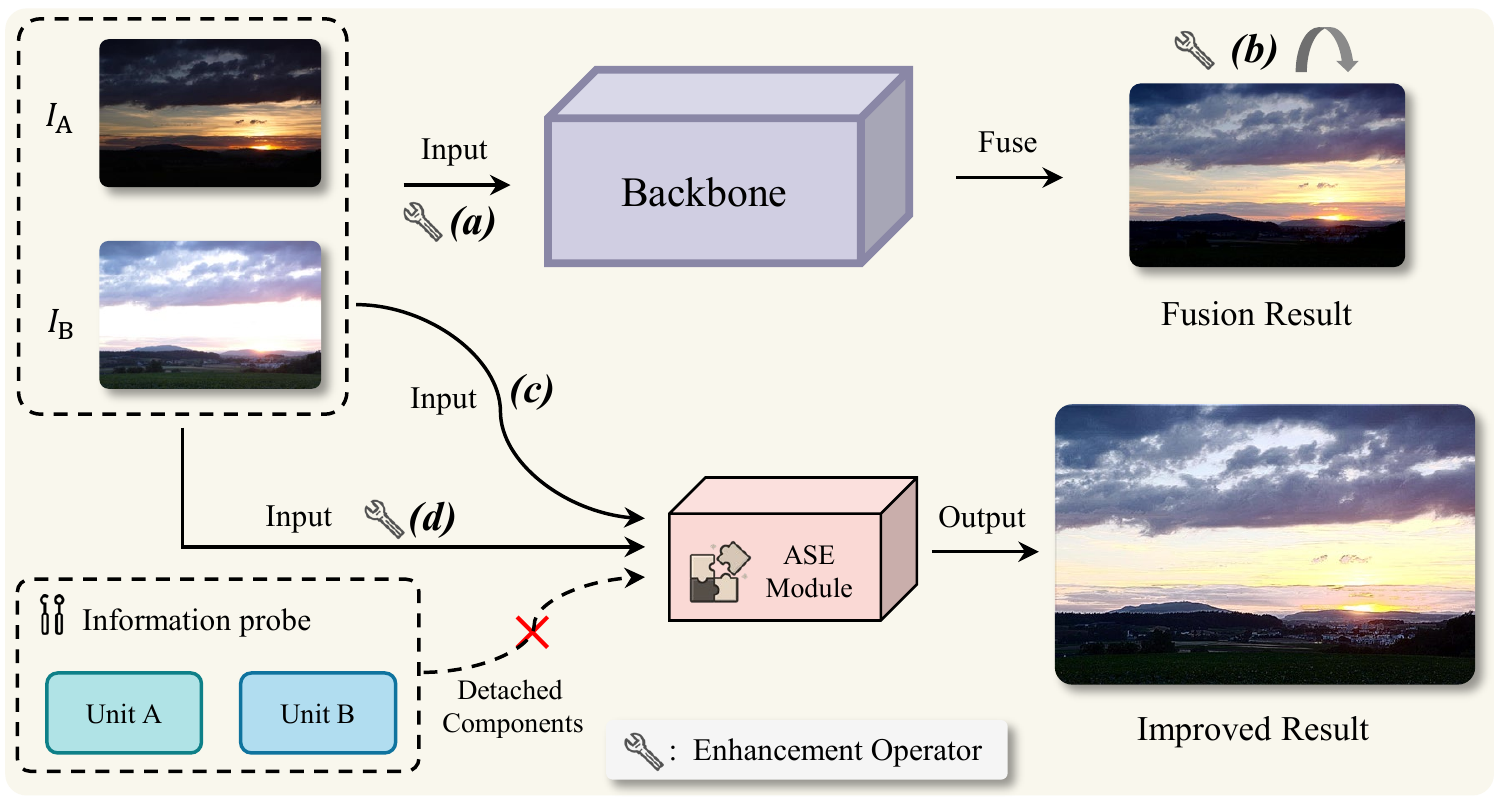}
\end{center}
   \caption{Illustration of different ablation experiments investigating the enhancement operation.
   The enhancement operator  corresponds to the sharpening operation in the booster layer.
   (a): Enhancing the input for the backbone method;
   (b): Directly enhancing the backbone method's fusion result;
   (c): Without the information probe, the input for the ASE module is the source images;
   (d): Without the information probe, the input for the ASE module is the enhanced source images.   
   }
   %($I_{\textrm{vis}}$: visible image; $I_{\textrm{ir}}$: infrared image)}
   %The degraded images Since the quality of the reconstructed images is linked to that of the fusion results. The beneficial disturbance contained in the enhanced source images can prompt the rebuilding of the high-quality fused images.}
\label{positionAblation}
\end{figure}

%\subsubsection{Qualitative experiments}
\subsubsection{Qualitative Experiments}
As a gradient-based image fusion method, U2Fusion delivers promising results in other fusion tasks.
In the case of the MEIF task, as shown in Fig.~\ref{DSCIE_qualitative}, its gradient-based information measurement ignores the adaptation of the exposure setting and the results tend to preserve more information from the underexposed image.
This observation indicates that it is tricky to consider all aspects of the fusion process in a single stage.
%This is led by the mixing of the imaging and fusion in one-stage.
%This is due to the inappropriate weighting issue the one-stage fusion methods.
By contrast, our booster effectively mitigates this issue by lighting the dark area of the original results in the second stage.
Meanwhile, as shown in the magnified regions, our booster also enhances the edge information and generates results of higher clarity.
When compared with other SOTA methods (SDNet and MUFusion), the enhanced fusion results benefit from maintaining an appropriate level of exposure and succeed in preserving abundant texture details.

\subsubsection{Quantitative Experiments}
We also compare the performance of different MEIF methods in the image quality assessments.
As shown in Table~\ref{table_quantative_MEIF}, the enhanced traditional and learning-based (U2Fusion) methods achieve consistent improvements in these metrics, as in the previous MFIF task.
Note that the best performance in most of the metrics over other SOTA methods in multiple fusion tasks demonstrates the powerful generalization capability of our concise booster design.
\begin{table*}[tbp]
  \centering
  \caption{Quantitative results of the ablation experiments in three different fusion tasks.}
  \label{tableMoreMFIFQuantitative}  
  \resizebox{\linewidth}{!}{
    \begin{tabular}{c|ccccc|ccccc|ccccc}
    \toprule
    \multirow{2}{*}{Methods} & \multicolumn{5}{c|}{IVIF (Backbone: GTF)} & \multicolumn{5}{c|}{MFIF (Backbone: CSR)} & \multicolumn{5}{c}{MEIF (Backbone: U2Fusion)} \\
\cmidrule{2-16}          & SD    & EN    & VIF   & EI    & Qabf  & SD    & EN    & VIF   & EI    & Qabf  & SD    & EN    & VIF   & EI    & Qabf \\
    \midrule
    Baseline & 50.16  & 7.35  & 0.58  & 44.13  & 0.45  & 54.09  & 7.16  & 1.09  & 75.04  & \textbf{0.72 } & 49.01  & 7.21  & 1.69  & 83.00  & \textbf{0.64 } \\
    Baseline + (a) & 28.99  & 6.64  & 0.23  & 33.25  & 0.33  & 31.21  & 6.59  & 0.50  & 68.26  & 0.62  & 30.30  & 6.70  & 0.83  & 74.14  & 0.53  \\
    Baseline + (b) & 24.35  & 6.40  & 0.18  & 31.51  & 0.30  & 29.19  & 6.44  & 0.46  & 65.88  & 0.62  & 27.50  & 6.50  & 0.69  & 65.99  & 0.53  \\
    Baseline + (c) & 50.97  & 7.37  & 0.60  & 46.68  & 0.41  & 58.91  & 7.26  & 1.28  & 78.47  & 0.67  & 51.39  & 7.27  & 2.14  & 113.03  & 0.54  \\
    Baseline + (d) & 51.79  & 7.39  & 0.72  & 70.74  & 0.41  & 60.74  & \textbf{7.37 } & 1.75  & 126.89  & 0.47  & 52.07  & 7.39  & 2.29  & \textbf{147.93 } & 0.39  \\
    \rowcolor[rgb]{ .851,  .851,  .851} Baseline (FB) & \textbf{53.26 } & \textbf{7.41 } & \textbf{0.88 } & \textbf{73.19 } & \textbf{0.48 } & \textbf{61.17 } & 7.33  & \textbf{1.79 } & \textbf{127.01 } & 0.45  & \textbf{58.57 } & \textbf{7.51 } & \textbf{2.51 } & 134.52  & 0.43  \\    
    \bottomrule
    \end{tabular}%
}
\end{table*}%

\subsection{Ablation Experiments}
\label{sectionAblation}
\subsubsection{The Impact of the Information Probe and the Booster Layer}
In this section, we present more ablation experiments evaluating our booster on different image fusion tasks.
An illustration of our experimental settings is presented in Fig.~\ref{positionAblation}.

Firstly, we validate the need to deploy the second stage to enhance the fusion results.
In setting (a), we enhance the source images for the backbone method to make it produce more promising results.
In setting (b), we directly enhance the fusion result of the backbone method.
On the other hand, we want to validate the effectiveness of the proposed information probe and the corresponding enhancement procedure used in the booster layer.
In settings (c) and (d), we discard the information probe and only use the source images (setting (c)) and enhanced source images (setting (d)) as the input of the ASE module, respectively.
%are used to demonstrate the rationality of using the specific information enhancement strategy in the booster layer between the information probe and the ASE module.
These four experiments are all conducted independently.
Here, the enhancement relates to the sharpening operation in the booster layer (the detail layer is from the image itself).

As shown in Table~\ref{tableMoreMFIFQuantitative}, our strategy in the booster layer enables the backbone methods to have the best overall performance in these three fusion tasks.
Specifically, the setting in our FusionBooster performs better in the IVIF task (5 best metrics).
As shown in Fig.~\ref{figureMoreAblationQualitative}, directly enhancing the source images or fusion results in the first stage will make the IVIF results blurred to a certain extent, and the bad exposure settings in the MEIF results cannot be improved (settings (a) and (b)).
Without using the information probe to divide different components apart (settings (c) and (d)), the enhancement quality cannot be guaranteed, \textit{e.g.}, the edge information from the IVIF result is not clear and the enhanced results from the MEIF task tend to produce some artifacts.

\begin{figure}[t]
\begin{center}
\includegraphics[width=1\linewidth]{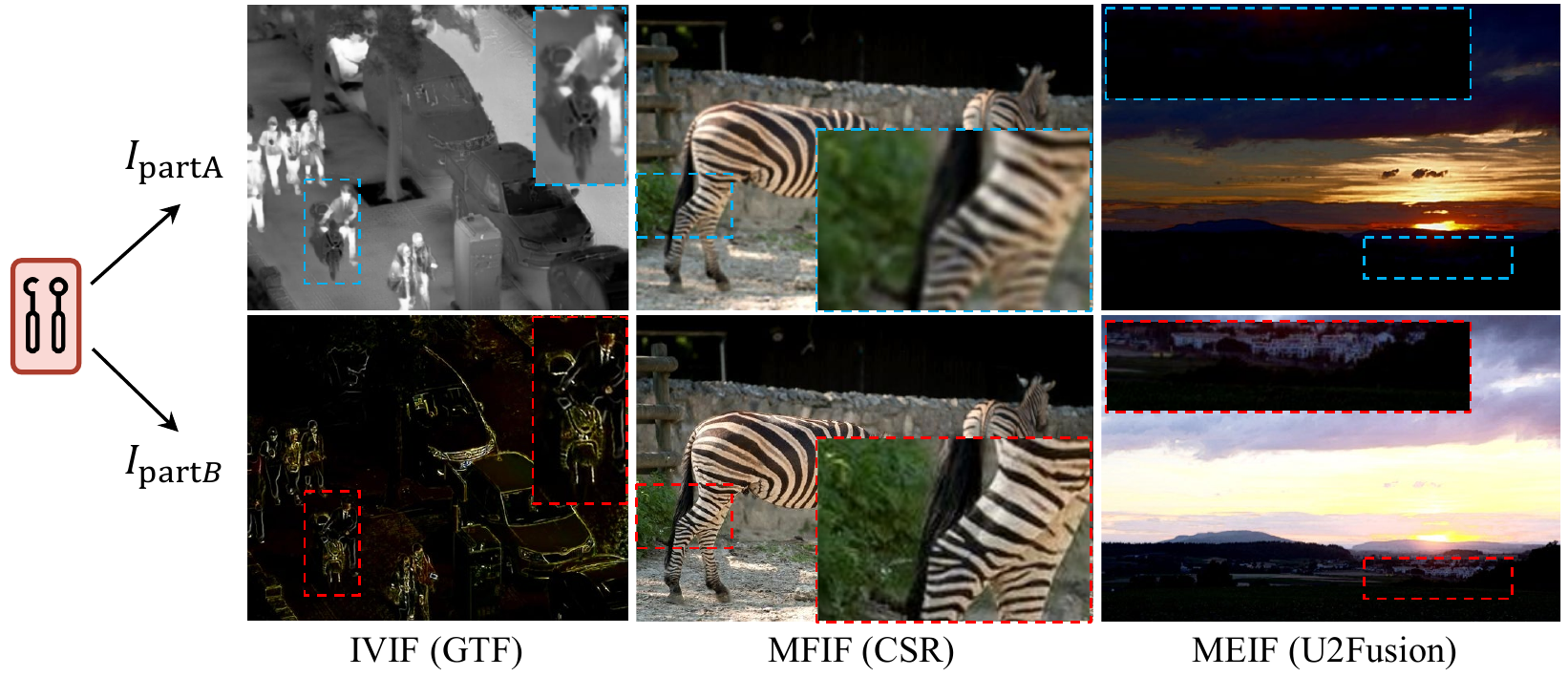}
\end{center}
   \caption{The perception results of the information probe otained in three image fusion tasks.
   As shown in the highlighted regions of the MFIF results, compared with the other two perception results, the information probe does not produce output with a focus on different areas of the source images.
   Instead, it only produces images with all-focused or all-blurred styles.
   }
   %($I_{\textrm{vis}}$: visible image; $I_{\textrm{ir}}$: infrared image)}
   %The degraded images Since the quality of the reconstructed images is linked to that of the fusion results. The beneficial disturbance contained in the enhanced source images can prompt the rebuilding of the high-quality fused images.}
\label{perceptionResults}
\end{figure}

\begin{figure}[t]
\begin{center}
\includegraphics[width=1\linewidth]{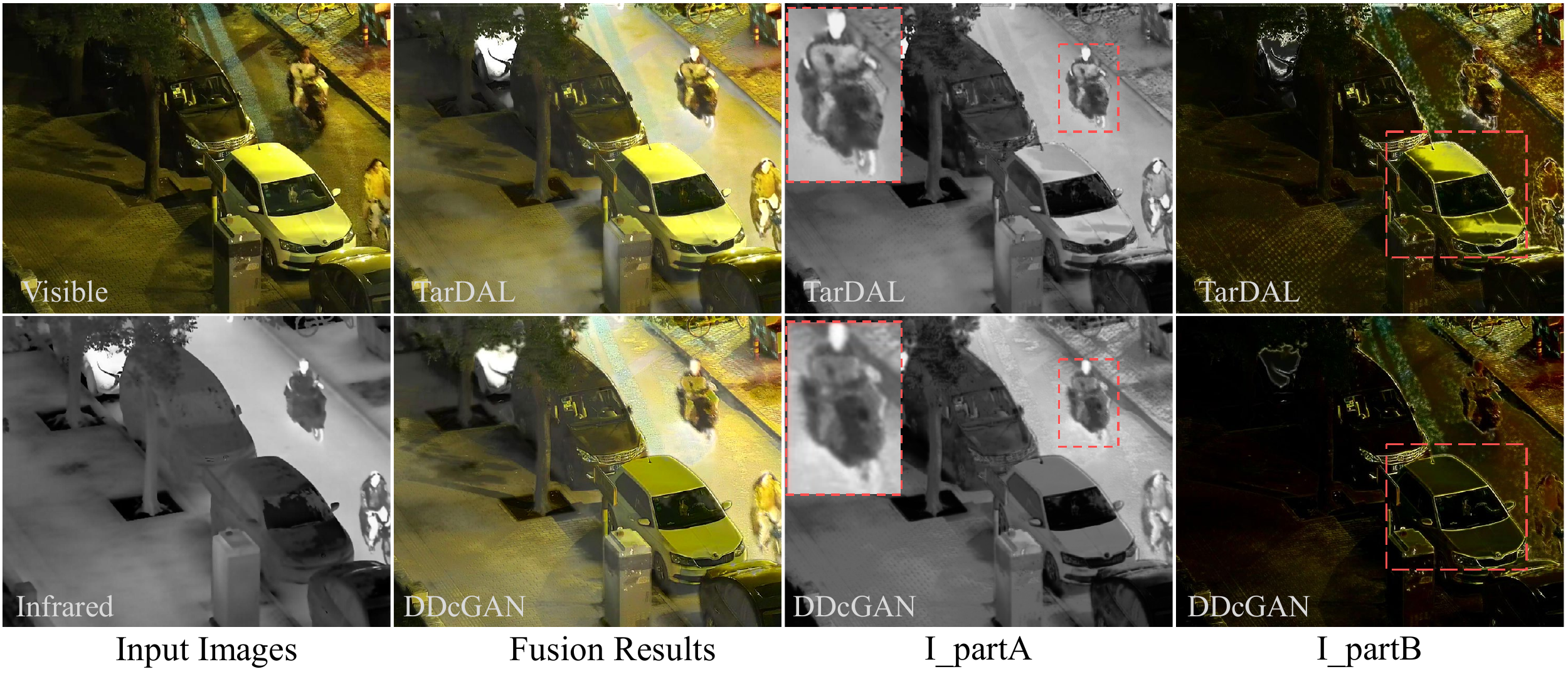}
\end{center}
   \caption{Reconstruction examples of the information probe oftained for two representative methods TarDAL and DDcGAN.
   As denoted by the red boxes, the degradation of the reconstructed images is related to the fusion performance of the initial results.
   }
   %($I_{\textrm{vis}}$: visible image; $I_{\textrm{ir}}$: infrared image)}
   %The degraded images Since the quality of the reconstructed images is linked to that of the fusion results. The beneficial disturbance contained in the enhanced source images can prompt the rebuilding of the high-quality fused images.}
\label{fig_degradationExamples}
\end{figure}

\subsubsection{An Analysis of the Information Probe}
\label{analysisInformationProbe}
The performance gain of our booster on the MFIF task is not as significant as it is for the other two tasks, \textit{i.e.}, the differences between the setting (d) and our booster are not very distinct in the visualization results.
We carried out further experiments to investigate this issue.
In Fig.~\ref{perceptionResults}, we show the perception results of our information probe in the case of these three fusion tasks.
In the IVIF task and the MEIF task, our booster can coarsely recover the complementary source images.
%as shown in the highlighted regions of the perception results on
However, in the MFIF task, lacking the necessary depth information, our probe can only produce images with either all-focused or all-blurred styles, which are not consistent with the source images.
In this particular example, the limited overlap between the disentangled components and the  supplementary source images leads to suboptimal enhancement effects.

The key aim of our FB is to enhance the components outputted by the information probe to improve the fusion performance.
The success of this operation is related to degradations caused by the information probe.
Specifically,  inaccurate estimates of weighting, and any artifacts injected by the backbone will make it difficult to reconstruct the source images from the initial fusion result.
In these circumstances the separated components will tend to degrade.
We argue that the quality of the fusion results is correlated with the degree of degradation of the reconstructed images.
Thus, if we recover these components successfully, and then combine them, theoretically, we should obtain better fused images.

From previous experiments, we notice that the TarDAL is able to produce fused images with a high-quality visual effect,
while the DDcGAN generates some noise and artifacts in their fusion results.
In Fig.~\ref{fig_degradationExamples}, we use these two methods to conduct experiments investigating the impact of degradation.
As shown in the red boxes, with higher image quality, the extent of degradation produced by TarDAL is less than that of the DDcGAN.
For example, the thermal radiation looks clearer in the infrared component of TarDAL.
Meanwhile, in the visible spectrum, the DDcGAN cannot recover the texture details from the image as effectively as TarDAL.
These observations are consistent with our expectations of the effect of degradation caused by the information probe.

\begin{figure*}[t]
\begin{center}
\includegraphics[width=1\linewidth]{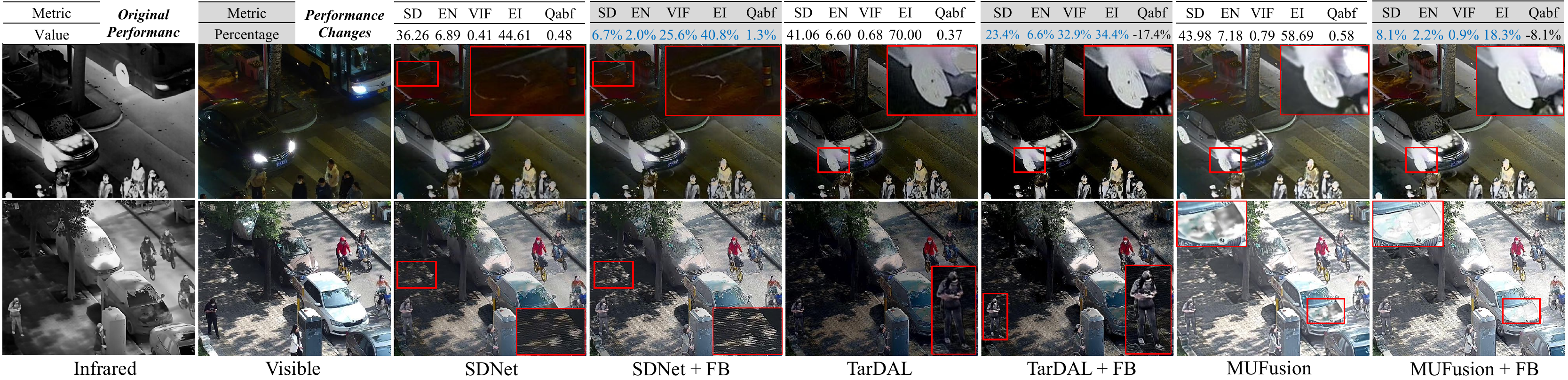}
\end{center}
   \caption{More results of the proposed FusionBooster combined with other advanced IVIF algorithms.
   As denoted in the highlighted regions, integrating SDNet with our booster preserves the detail information from the background better. The salience of the targets is increased in the results of TarDAL, and the artifacts contained in the MUFusion are eliminated.
   ( \textcolor[rgb]{0, 0, 0}{\textcolor[rgb]{ 0,  .439,  .753}{\textbf{Blue}}}: Better performance obtained using FusionBooster )
   }
   %($I_{\textrm{vis}}$: visible image; $I_{\textrm{ir}}$: infrared image)}
   %The degraded images Since the quality of the reconstructed images is linked to that of the fusion results. The beneficial disturbance contained in the enhanced source images can prompt the rebuilding of the high-quality fused images.}
\label{figureMoreQualitativeIVIF}
\end{figure*}

\begin{figure*}[t]
\begin{center}
\includegraphics[width=1\linewidth]{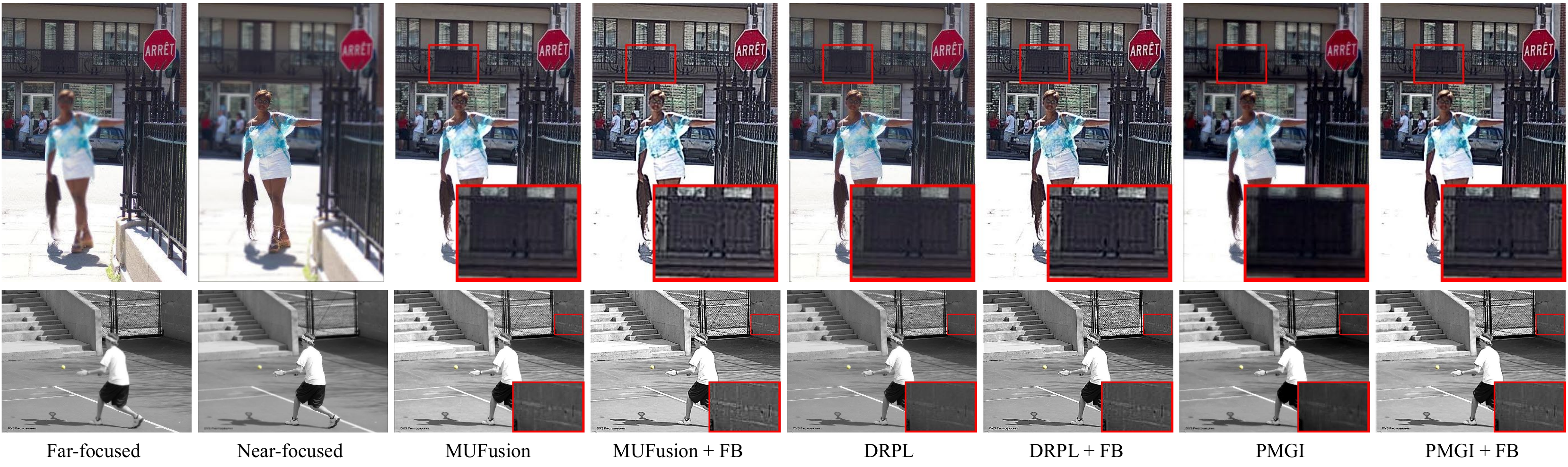}
\end{center}
   \caption{More qualitative results of the proposed FusionBooster integrated with other MFIF algorithms.
   As denoted in the magnified areas, our booster, integrated with the advanced fusion methods MUFusion and DRP, presents clearer texture details.
   The original PMGI produces blurring.
   Since the high-quality supplement source is used in the booster layer, our FusionBooster effectively mitigates this issue.
   }
   %($I_{\textrm{vis}}$: visible image; $I_{\textrm{ir}}$: infrared image)}
   %The degraded images Since the quality of the reconstructed images is linked to that of the fusion results. The beneficial disturbance contained in the enhanced source images can prompt the rebuilding of the high-quality fused images.}
\label{figureMoreQualitativeMFIF}
\end{figure*}

\begin{figure*}[t]
\begin{center}
\includegraphics[width=1\linewidth]{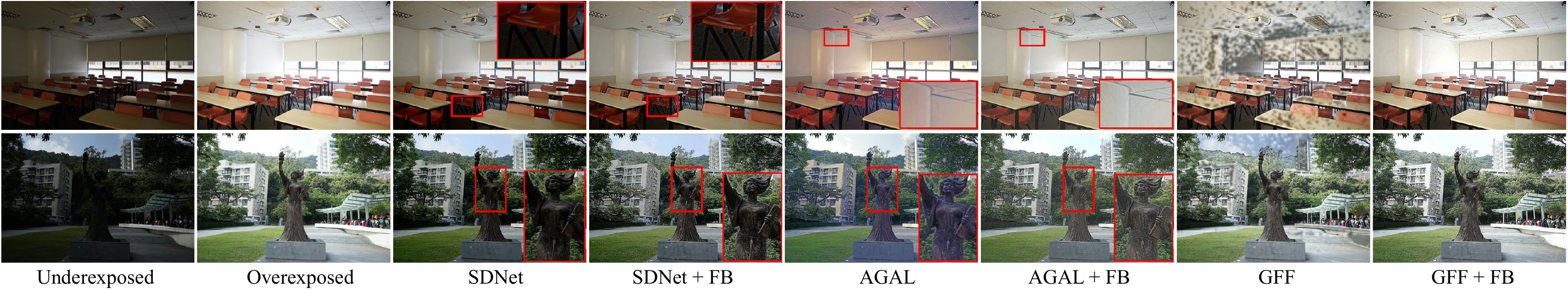}
\end{center}
   \caption{More qualitative results obtained by the proposed FusionBooster in conjunction with with other MEIF algorithms.
   As shown in the highlighted regions, SDNet obtains better exposure in the dark regions thanks to FusionBooster.
   The unnatural colour and artifacts visible in the fusion results of AGAL and GFF are also effectively removed in the refined output.
   }
   %($I_{\textrm{vis}}$: visible image; $I_{\textrm{ir}}$: infrared image)}
   %The degraded images Since the quality of the reconstructed images is linked to that of the fusion results. The beneficial disturbance contained in the enhanced source images can prompt the rebuilding of the high-quality fused images.}
\label{figureMoreQualitativeMEIF}
\end{figure*}

\begin{figure}[t]
\begin{center}
\includegraphics[width=1\linewidth]{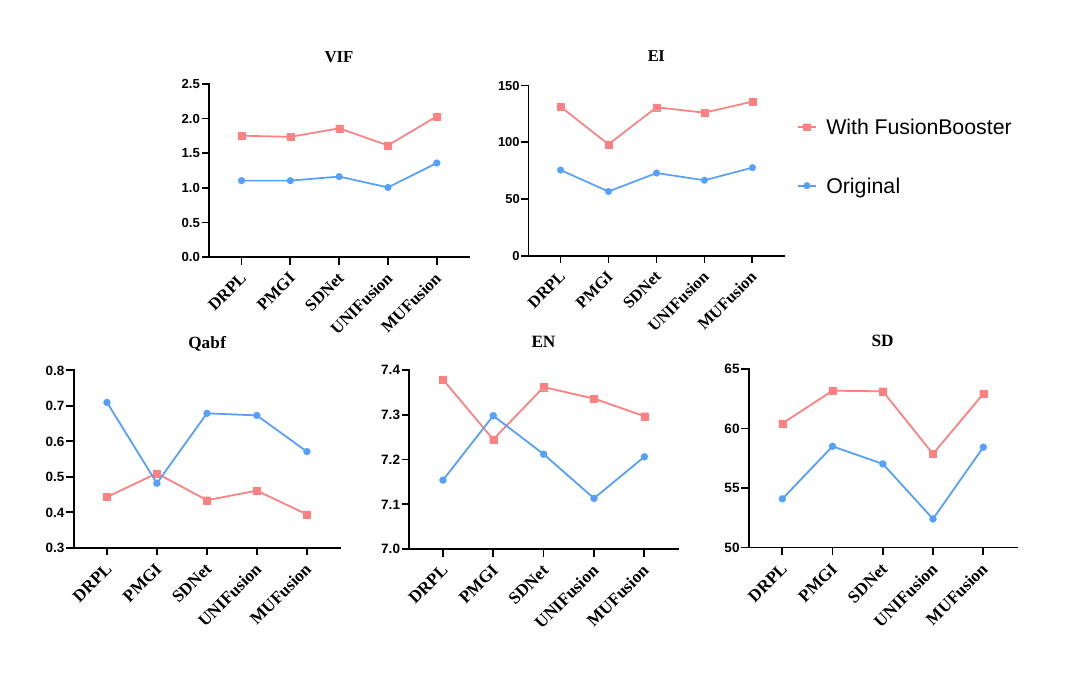}
\end{center}
   \caption{The quantitative results of the proposed FB combined with more MFIF algorithms.
   Generally, our FB can consistently enhance the performance of most image fusion methods.
   However, the gain from the FB can be very different for various backbone methods (the performance of Qabf and EN).
   }
\label{figureMoreQuantitativeMFIF}
\end{figure}

\begin{table}[tbp]
  \centering
    \caption{The quantitative results obtained by the proposed FusionBooster (FB) in conjunction with more MEIF algorithms. ( \textcolor[rgb]{0, 0, 0}{\textcolor[rgb]{ 0, 0,  0}{\textbf{Bold}}}: Better performance obtained using  FusionBooster )}
  \label{tableMoreMEIFQuantitative} 
  \resizebox{0.8\linewidth}{!}{  
    \begin{tabular}{cccccc}
    \toprule
    Method & SD    & EN    & VIF   & Qabf  & EI \\
    \midrule
    GFF   & 47.072  & 7.382  & 1.527  & 0.653  & 84.293  \\
    +FB   & \textcolor[rgb]{ 0, 0,  0}{\textbf{53.557 }} & 7.132  & \textcolor[rgb]{ 0, 0,  0}{\textbf{2.002 }} & 0.514  & \textcolor[rgb]{ 0, 0,  0}{\textbf{108.172 }} \\
    \midrule
    \midrule
    DeppFuse & 46.091  & 7.156  & 1.295  & 0.696  & 60.527  \\
    +FB   & 45.918  & \textcolor[rgb]{ 0, 0,  0}{\textbf{7.169 }} & \textcolor[rgb]{ 0, 0,  0}{\textbf{1.320 }} & 0.590  & \textcolor[rgb]{ 0, 0,  0}{\textbf{80.146 }} \\
    \midrule
    \midrule
    SDNet & 44.135  & 7.035  & 1.299  & 0.677  & 72.497  \\
    +FB   & 43.890  & \textcolor[rgb]{ 0, 0,  0}{\textbf{7.183 }} & \textcolor[rgb]{ 0, 0,  0}{\textbf{1.408 }} & 0.521  & \textcolor[rgb]{ 0, 0,  0}{\textbf{101.926 }} \\
    \midrule
    \midrule
    AGAL  & 43.725  & 7.106  & 1.314  & 0.655  & 71.954  \\
    +FB   & \textcolor[rgb]{ 0, 0,  0}{\textbf{44.930 }} & \textcolor[rgb]{ 0, 0,  0}{\textbf{7.208 }} & \textcolor[rgb]{ 0, 0,  0}{\textbf{1.383 }} & 0.507  & \textcolor[rgb]{ 0, 0,  0}{\textbf{97.936 }} \\
    \midrule
    \midrule
    MUFusion & 49.682  & 7.231  & 1.637  & 0.716  & 70.179  \\
    +FB   & \textcolor[rgb]{ 0, 0,  0}{\textbf{49.836 }} & \textcolor[rgb]{ 0, 0,  0}{\textbf{7.293 }} & \textcolor[rgb]{ 0, 0,  0}{\textbf{1.694 }} & 0.561  & \textcolor[rgb]{ 0, 0,  0}{\textbf{96.542 }} \\
    \bottomrule
    \bottomrule
    \end{tabular}%
}
\end{table}%

\subsection{More Results in Different Fusion Tasks}
\subsubsection{More Results in the IVIF Task}
\label{more_IVIF}
In this section, we present more experimental results of our booster integrated with other IVIF algorithms.
In this experiment, both day and night scenes are covered in our test images.
As shown in Fig.~\ref{figureMoreQualitativeIVIF}, our booster consistently strengthens the performance of these algorithms in  the following sense: the preservation of more texture details in the background regions (SDNet), the improvement in the capture of salience of the thermal radiation (TarDAL++), and the reduction of  artifacts (MUFusion).
These advanced methods all achieve better quantitative performance when combined with our booster in different image quality assessments, \textit{e.g.}, huge increase in the performance of VIF and EI, which demonstrates the superiority of our booster in the performance of  the IVIF task.
In some of the cases, \eg, the fusion results of MUFusion(FB), our booster enhances the performance of the quantitative results only marginally.
However, the benefit of reducing artifacts should not be underestimated, which is not well reflected by the adopted metrics.
%as the original imaging quality of this method is not very promising.

%%%%%%%%% BODY TEXT
\subsubsection{More Results in the MFIF Task}
We conducted further experiments relating to the MFIF task to validate the effectiveness of the proposed FusionBooster.
As shown in the highlighted regions of Fig.~\ref{figureMoreQualitativeMFIF}, the blurring issue in the fusion result of PMGI~\cite{zhang2020rethinking} is effectively mitigated by our booster.
The proposed FB also enhances the preserved detail information of other fusion approaches, \textit{e.g.}, the fence in front of the door is clearer in the refined images.
In Fig.~\ref{figureMoreQuantitativeMFIF}, we also evaluate the quantitative performance of different image fusion methods and their upgraded versions.
Our FusionBooster effectively boosts the performance of different fusion methods on most of these five metrics, \textit{e.g.}, around 33\% improvement in the metric of visual information fidelity for the MUFusion.
Although the PMGI has worse performance in the metric of EN, the higher quality images yielded in the qualitative experiments indicate that our booster works well in conjunction with this algorithm.
Besides, the much more abundant texture details from the improved PMGI is also consistent with the significant increase in the metric of EI. 

\subsubsection{More Results in the MEIF Task}
In this section, we conduct experiments to evaluate FusionBooster integrated with more MEIF algorithms.
%on the MEIF task.
As shown in Fig.~\ref{figureMoreQualitativeMEIF},  applying our booster to the SDNet~\cite{zhang2021sdnet} results in a more appropriate exposure in the dark regions of the original fused images.
Note also, as presented in the magnified regions, the original results of the AGAL~\cite{liu2022AGAL_ME} and the traditional approach GFF~\cite{li2013gff} are not satisfactory.
Specifically, they exhibit unnatural colour and artifacts in the output images.
The refined results  appear to address these issues effectively.

Finally, for the quantitative experiments (Table~\ref{tableMoreMEIFQuantitative}), the reasons for poor performance on the metric of Qabf have been explained in Section.~\ref{SectionMFIFQuantitative}.
In consistency with the conclusions drawn from the previous experiments, our booster enables all MEIF methods to achieve significant improvements in most of the image quality assessments.

\begin{table*}[tbp]
  \centering
  \caption{The inference time and the model size comparison of different methods on 250 image pairs from the LLVIP dataset. (\textcolor[rgb]{ 0,  0,  0}{\textbf{Bold}}: extra cost)}
  \label{tableEfficiency}%
  \vspace{2mm}
  \resizebox{\linewidth}{!}{
    \begin{tabular}{cccccc|cc|cc|cc}
    \hline
    Metric & U2Fusion & MUFusion & YDTR  & SeAFusion & DenseFuse & GTF   & GTF + FB  & DDcGAN & DDcGAN + FB & SDNet & SDNet + FB \\
    \hline
    Time (s) & 66.90  & 52.98  & 28.01  & 7.51  & 1.82  & 128.67  & 130.66 (\textcolor[rgb]{ 0,  0,  0}{\textbf{+1.99}}) & 123.12  & 125.04 (\textcolor[rgb]{ 0,  0,  0}{\textbf{+1.92}}) & 2.14  & 4.13 (\textcolor[rgb]{ 0,  0,  0}{\textbf{+1.99}}) \\
    Model size (MB) & 2.51  & 2.12  & 0.85  & 0.65  & 0.29  & --    & --    & 21.18  & 21.35 (\textcolor[rgb]{ 0,  0,  0}{\textbf{+0.17}}) & 0.26  & 0.43 (\textcolor[rgb]{ 0,  0,  0}{\textbf{+0.17}}) \\
    \hline
    \end{tabular}%
    }
\end{table*}%

\subsection{Comparison of the Computational Complexity and Model Size}
In this section, we provide the statistics of additional time consumption and model size burden for several image fusion methods  utilizing the proposed FusionBooster.
The information is presented  in Table~\ref{tableEfficiency}, where we collect the inference time of several approaches, as well as their model sizes in the context of the IVIF task on the LLVIP dataset.
While achieving much better performance in various fusion tasks, our booster increases the time consumption of the baseline methods by only around 2 seconds on 250 infrared and visible image pairs, and increases the size of the model by less than 200KB.
Such lightweight solution offers attractive advantages in comparison with the expensive computational cost issue of the existing enhancement-based methods.
%Besides, our model size is also very tiny, only 0.17 MB extra 

\section{Conclusion}
In this paper, we proposed an image fusion enhancer based on a divide and conquer strategy guided by an innovative information probe. It is the first time such a universally applicable boosting paradigm is proposed in the literature.
%Specifically, it is tricky to obtain ideal image fusion results in the current image fusion paradigm.
Given a fused image from an arbitrary method, \textit{e.g.}, an IVIF algorithm, we first decompose the initial result into different components. The information probe gauges the affinity of the components to the input images, and filters them to yield an improved fused image. The difference signal iteratively drives the update of the FusionBoster parameters. 
In this way, we effectively mitigate the information loss and image blurring issues in the backbone.
%, and consequently produce fused images with sharper edge information.
%After that, we conquer and fix these two parts by utilizing the booster layer.
%Finally, we resort to the assembling module to yield the final output.
%Our two-stage fusion paradigm is designed based on the intrinsic attribute of the fused image, \textit{i.e.}, the preservation of information amount in the fused images.
The nested AE design of the network architecture and the loss function are the key ingredients of the improved performance at the expense of a minor increase in the computational cost.
Compared with other extra modules required by the enhancement-based methods, the proposed booster can be applied to different fusion tasks very effectively.
Moreover, it significantly boosts various fusion approaches, including the traditional and learning-based methods.

Although our booster  significantly enhances existing image fusion methods,
it leaves some scope for future research. 
%\textcolor{red}{I do not understand what you are trying to say in the rest of the paragraph} 
Firstly, in our FB, we study the image fusion enhancement only from the perspective of information retention.
As presented in some of the experiments, our FB cannot always improve the image fusion performance.
Thus, investigating diverse manners to disentangle and analyse the fused images may probably benefit the performance of the booster.
%Thus, the incompatibility between the inner design and these operations may hurt the original performance.
%Such design cannot always effectively enhance the fusion performance, as presented in some of the experimental results.
%Actually, it may also hurt the performance of
%being an additional module for existing image fusion methods, the proposed solution is not conceived as a new image fusion method in this paper.
%Thus, when compared with our method, the limited manner is to replace the corresponding enhancement model with our FB or using an advanced method upgraded by our FB as the competitor.
Secondly, the effective enhancement strategy delivered by the booster layer could potentially be further improved in the future by a trainable booster network.
Finally, the proposed booster has only been validated in a limited number of fusion tasks. The generalization ability of this approach remains to be proven in other applications.

\section*{Availability of Data and Materials}
Information on access to the datasets supporting the conclusions of this article is included therein.

\section*{Competing Interests}
The authors declare that they have no conflict of interest.

\begin{acknowledgements}
This work was supported by the National Natural Science Foundation of China (62020106012, U1836218, 62106089, 62202205), the 111 Project of Ministry of Education of China (B12018), the Engineering and Physical Sciences Research Council (EPSRC) (EP/R018456/1, EP/V002856/1), and the Postgraduate Research \& Practice Innovation Program of Jiangsu Province (KYCX23\_2525).
\end{acknowledgements}

\bibliographystyle{spbasic}      % basic style, author-year citations
\bibliography{ref.bib}

% % Non-BibTeX users please use
% \begin{thebibliography}{}
% %
% % and use \bibitem to create references. Consult the Instructions
% % for authors for reference list style.
% % Format for Journal Reference
% Author, Article title, Journal, Volume, page numbers (year)
% Author, Book title, page numbers. Publisher, place (year)
% \end{thebibliography}

\end{document}